\RequirePackage{fix-cm}
\documentclass[journal]{IEEEtran}
\usepackage{graphicx} 
\usepackage{amsmath} 
\usepackage{amssymb}  
\usepackage{algorithm}
\usepackage{gensymb}
\usepackage[caption=false]{subfig}
\usepackage{multirow}
\usepackage[noend]{algpseudocode}
\usepackage{tikz}
\usetikzlibrary{fit,positioning}

\begin{document}

\title{Multi-Modal Active Perception for\\Information Gathering in Science Missions}
        
\author{Akash~Arora,
		P.~Michael~Furlong,
		Robert~Fitch,
        Salah~Sukkarieh,
        and~Terrence~Fong,

\thanks{This work was supported by the Australian Center for Field Robotics and NASA Mojave Volatiles Prospector~(MVP) project in the NASA Moon and Mars Analog Missions Activities~(MMAMA) Program}
\thanks{A.~Arora, R.~Fitch, and S.~Sukkarieh are with the Australian Centre for Field Robotics, The University of Sydney, NSW, Australia. (Email:~aaro4138@uni.sydney.edu.au; salah.sukkarieh@sydney.edu.au)}
\thanks{R.~Fitch is also with the Center for Autonomous Systems, University of Technology Sydney, Sydney, Australia. (Email: rfitch@uts.edu.au)}
\thanks{P.~M.~Furlong and T.~Fong are with the Intelligent Robotics Group, NASA Ames Research Center, CA, USA. (Email:~padraig.m.furlong@nasa.gov; terry.fong@nasa.gov)}
}
\maketitle

\begin{abstract}
Robotic science missions in remote environments, such as deep ocean and outer space, can involve studying phenomena that cannot directly be observed using on-board sensors but must be deduced by combining measurements of correlated variables with domain knowledge. Traditionally, in such missions, robots passively gather data along prescribed paths, while inference, path planning, and other high level decision making is largely performed by a supervisory science team. However, communication constraints hinder these processes, and hence the rate of scientific progress. This paper presents an active perception approach that aims to reduce robots' reliance on human supervision and improve science productivity by encoding scientists' domain knowledge and decision making process on-board. We use Bayesian networks to compactly model critical aspects of scientific knowledge while remaining robust to observation and modeling uncertainty. We then formulate path planning and sensor scheduling as an information gain maximization problem, and propose a sampling-based solution based on Monte Carlo tree search to plan informative sensing actions which exploit the knowledge encoded in the network. The computational complexity of our framework does not grow with the number of observations taken and allows long horizon planning in an anytime manner, making it highly applicable to field robotics. Simulation results show statistically significant performance improvements over baseline methods, and we validate the practicality of our approach through both hardware experiments and simulated experiments with field data gathered during the NASA Mojave Volatiles Prospector science expedition. 

\end{abstract}

\section{Introduction}
\label{sec::introduction}
Information gathering using mobile robots in dangerous or hard-to-access environments has significantly improved humanity's ability to understand the Solar System~\cite{grotzinger2012mars}. In such missions, scientists can be interested in variables or phenomena that cannot be directly measured using on-board sensors, but instead must be inferred through proxy measurements--measurements that are correlated with the variables of interest. Examples include mapping water abundance in large environments by measuring neutron flux~\cite{heldmann2015mojave}, inferring the health of aquatic life by monitoring chemical concentrations~\cite{dunbabin2012robots}, and searching for evidence of life on Mars through detection of various biomarkers and organic compounds~\cite{leshin2013volatile}.  

\begin{figure*}[ht]
\centering
	\hspace*{\fill}%
    \subfloat[b][]{
    \centering
    	\includegraphics[trim = {0 0 1.5cm 0},clip,height=4cm]{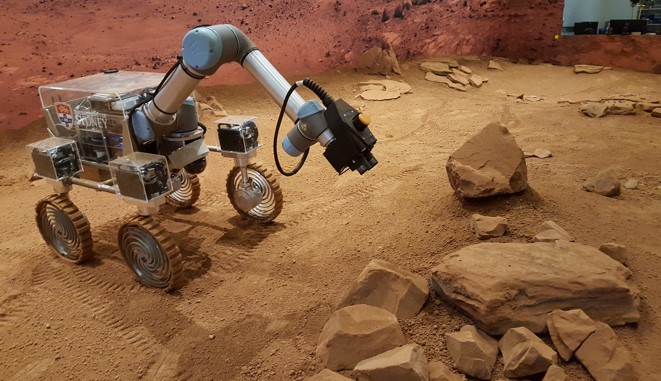}}
    ~ 
    \hspace*{\fill}%
    \subfloat[b][]{
    \centering
        \includegraphics[trim = {1.5cm 0 0 0},clip,height=4cm]{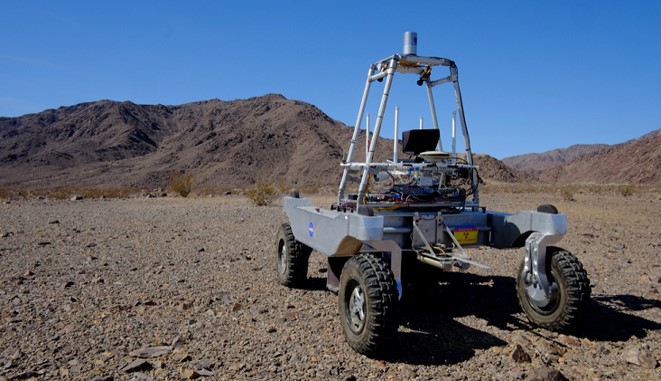}}
    \hspace*{\fill} %
    \hfill
    \subfloat[b][]{
    \centering
        \includegraphics[trim = {1cm 0 0 0},clip,height=4cm]{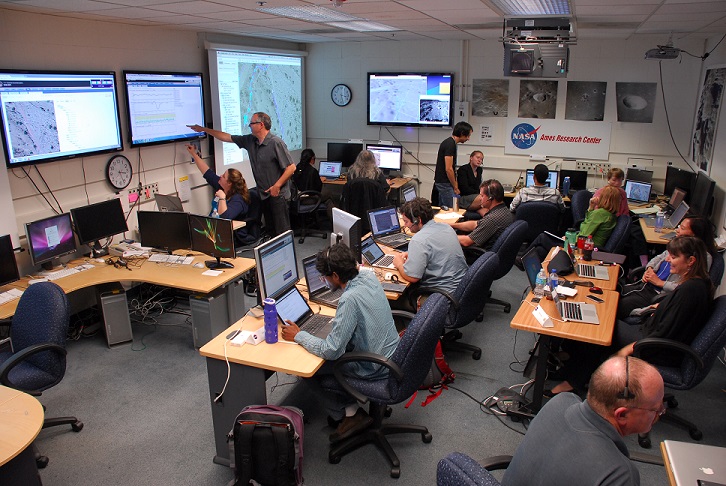}}
    \hspace*{\fill}%
\caption{The active information gathering missions we study in this paper. (a) The Continuum rover in a Mars-analog environment using its robotic arm camera to closely examine rocks; (b) NASA's KREX2 rover in the Mojave Desert mapping water abundance in the soil with a Neutron Spectrometer System; (c) The MVP mission team analyzing sensing data remotely and planning informative trajectories for the robot. \textbf{Photo credit for Figs. b-c: NASA Ames}}
\label{fig:continuum}
\end{figure*}

In science missions, like the upcoming Resource Prospector mission~\cite{andrews2014introducing}, robots largely act as mobile sensing platforms that passively collect data along prescribed paths, and pass it on to a supervising science team at the earliest opportunity. The science team analyzes the information, and combines it with past observations and domain knowledge to make inferences about the variables of interest. A cyclic process of data gathering, inference, and path planning is repeated until the science goals of the mission have been achieved.

However, robotic science missions often take place under strong communication constraints which limit how effectively this ``scientist in the loop" approach can be carried out. Latency delays may make direct teleoperation infeasible, while bandwidth limitations mean scientists must reason with incomplete information. Communication windows can also be short and sparse which means exploration plans cannot be rapidly updated in response to sensor data collected by robots. In this paper, we propose an active perception~\cite{bajcsy1988active} approach to information gathering in science missions, which aims to overcome the reliance on human supervision and increase science productivity by approximating scientists' inference and decision making processes on-board the robot. 

We consider two complications of active information gathering that existing approaches do not address. Firstly, in some missions, the variables of interest can be latent. Missions may involve studying abstract concepts, such as evidence of ancient riverbeds~\cite{grotzinger2012mars}, which require combining proxy measurements with scientific domain knowledge and information that is outside the robots' on-board sensory capabilities to infer. Secondly, robots can be equipped with heterogeneous sensing modalities which measure different subsets of environmental features but that also have different energy or time costs associated with their usage. When these costs are significant, such as when sensors require collecting subsurface samples with a drill, the robot has to be selective in where it deploys which sensor. 

Previous approaches attempt to bring active behavior to parts of the information gathering pipeline by allowing the robot to divert off preplanned paths and take additional measurements with various instruments if it detects regions in the environment whose visual features match some criteria~\cite{castano2007oasis,estlin2012aegis}. However, without a formal mapping between sensor observations and variables of interest encoded on-board, and access to suitable non-myopic planners, these methods do not gather information in a resource efficient manner. Recently, data driven approaches have become popular, especially in environmental monitoring applications~\cite{dunbabin2012robots}. Approaches combining Gaussian processes with uncertainty driven planning known as informative path planning have been particularly successful~\cite{ma2017data,das2013hierarchical}. However, finding paths that maximize information on a latent variable, and the additional decision of selecting \textbf{which} sensor to use in addition to planning informative waypoints is outside the scope of existing informative path planning algorithms~\cite{hollinger2014sampling,marchant2014bayesian,bender2013autonomous}.

In this paper, we study a new planning problem called \textit{multi-modal scientific information gathering}. Here a robot, equipped with heterogeneous sensing modalities, is required to plan paths and determine where to use which sensor to maximize the information gained about some variable of scientific interest (which may not be directly observable) while adhering to a resource budget. Our initial solution algorithm addresses this as two separate subproblems. The first is to create a mapping between sensor data and variables of interest, by encoding mission critical aspects of scientists' domain knowledge in a way the robot can reason about efficiently. The second subproblem is to plan paths and sensing actions that maximize the information gained on the variable of interest. 

The first subproblem involves designing a mapping representation which allows intuitive encoding of the scientists' domain knowledge regarding the variables of mission interest. The representation must fulfill three requirements. Firstly, it must be sufficiently expressive. Scientific knowledge is primarily composed of causal knowledge such as conditional dependencies between variables, class hierarchies, and mathematical or process models~\cite{hunter2010survey}. The representation must be suitable for capturing such causal relations. Secondly, in field environments, sensing is noisy, and the likelihood of errors in prior models is high. The representation must be robust to uncertainty. Lastly, since we are interested in online execution, the representation must also allow efficient inference. 

This paper utilizes Bayesian networks~(BNs) to create a mapping between sensor data and the variables of interest, while jointly and compactly modeling critical aspects of scientific knowledge, as well as other prior knowledge such as classifiers and sensor noise models. The BNs directed structure is well suited for modeling causal relationships, while the conditional probability parameters allow us to handle noisy observations and incomplete information robustly. There are also many approximate algorithms available for inference that make BNs well suited for field robots with limited computational power and real time mission constraints~\cite{pearl2014probabilistic}. 

The second subproblem is planning where to go, and deciding which sensor to deploy to learn about the variables of interest in a resource efficient manner. This requires reasoning about the scientific knowledge representation, mission constraints, and the perceived state of the environment. We cast this decision making process as an information gain maximization problem. Finding optimal solutions in partially observable environments is in general intractable~\cite{roy2005finding}. Optimizing information gain with respect to latent variables, and the additional planning dimension of deciding when to activate which sensor, breaks the assumptions of existing approximate informative path planning algorithms, e.g. ~\cite{hollinger2014sampling,marchant2014bayesian}.

In this paper we adapt Monte Carlo Tree Search~(MCTS) for approximate planning. MCTS techniques have led to state of the art performance in game-playing literature~\cite{silver2016mastering} and recently gained popularity in robotic applications~\cite{best2016decentralised,nguyen2016real} due to their ability to plan long horizons in an anytime manner. We adapt MCTS techniques to reason about domain knowledge and partially observable environments to generate non-myopic plans for multiple on-board sensing modalities in a principled manner while remaining anytime.

The proposed active perception approach, composed of a novel fusion of BN knowledge representation and MCTS approximate planning, extends the capabilities of information gathering robots in several significant ways. Firstly, robots can efficiently generate non-myopic informative sensing sequences with heterogeneous sensing modalities. Secondly, since mission relevant domain knowledge in encoded on-board, robots can do inference on latent variables they previously would not have been aware of. These new capabilities allows robots to directly gather information on abstract variables of interest in a resource efficient manner without having to rely on a supervisory science team to make inferences from observations and update sensing plans. The ability to reason about scientific domain knowledge in a principled manner also improves situational awareness of robots, which allows scientists to interact with the robot at higher levels of abstraction, and acts as a foundation for task level autonomy systems to be developed. These properties should lead to significant increases in science output of missions, and enable the possibility of exploring even more remote environments. 

Our active approach is illustrated through two mission scenarios. The first is a Mars exploration mission where the rover is required to infer the distribution of the types of environment (i.e. desert, volcanic or riverbed) in the map by observing geological features. The robot, equipped with two sensors, a camera and an idealized spectrometer, must decide where to move and which sensor to use at each time step while satisfying some sensing budget. We present extensive simulation results where our method outperforms baseline methods in terms of information gain and classification accuracy. We then demonstrate the practicality of our approach in an analog Martian environment where our experimental rover, Continuum (shown in Fig.~\ref{fig:continuum}a), plans and executes a science exploration mission in an end to end manner.  

In remote science missions, an accurate model of scientific knowledge is unlikely to exist prior to the mission. Often, the purpose of robotic exploration missions is to gain new knowledge in the first place. Our second case study is modeled on the Mojave Volatiles Prospector (MVP) project conducted by NASA Ames Research Center in the Mojave Desert in 2014~\cite{heldmann2015mojave}. The purpose of the MVP project was to test high tempo remote operations while attempting to estimate abundance of subsurface water using a Neutron Spectrometer System~(NSS) mounted on KRex2 pictured in Fig.~\ref{fig:continuum}b. At the end of the project, the sensor data was analyzed and a previously unknown relationship was discovered between the visual properties of terrain and the NSS readings, an example of incomplete prior knowledge~\cite{foil2016}. We show how we can recover from inaccuracies in scientific knowledge by modeling parameter uncertainties as random variables which are automatically updated and refined as observations are collected during the mission. We validate our approach with real data gathered during the MVP mission.



The main contributions of this paper are a formulation of the multi-modal scientific information gathering problem, a powerful initial solution algorithm based on BNs and MCTS, and lastly, extensive evaluation in simulation, on an experimental Mars rover, and using field data from a previous scientific expedition. We unify the key ideas introduced in conference versions of this work~\cite{arorairos,aroraFSR} and extend it in several ways. Firstly, a more general and detailed formulation and solution of the planning problem is provided with new analysis on worst case complexity. Secondly, the Mars exploration simulation experiments from \cite{arorairos} were rerun in more controlled settings which allowed measurement of statistical significance. Lastly, the practical hurdles of implementing a high level decision making algorithm on a real robot are more deeply discussed, and additional results are provided that characterize the effect of prior knowledge on algorithm performance. 

\section{Related Work}
\label{sec::relatedwork}
While the key ideas discussed in this paper apply to general robotic science missions, work most similar to ours has appeared in space robotics science autonomy research. We begin by reviewing the key literature in this area, followed by a discussion of knowledge representation and reasoning approaches in AI. We conclude with an overview on informative path planning algorithms. 

\subsection{Science autonomy in space robotics}
Mars rovers are subject to severe communication constraints, including low bandwidth, high latency and sparse communication windows that provide only one or two communication opportunities per day. A ``robotic astrobiologist" that understands data on-board and explores the environment with intelligence comparable to a human scientist during communication downtimes is an active area of research~\cite{ellery2017robotic}. We briefly discuss three major projects and how they differ from our work:  Carnegie Mellon's Life in Atacama project~\cite{cabrol2007life}, NASA's Onboard Autonomous Science Investigation System~(OASIS)~\cite{castano2007oasis} and European Space Agency's autonomy architecture for the planned ExoMars mission~\cite{van2005development}. 

\subsubsection{Life in Atacama project} 
\label{subsec:lifeatacama}
Approaches most similar to ours have resulted from CMU's Life in Atacama project. Smith et al. studied a scenario where the robot had to map the likelihood of the presence of living organisms in a grid world environment by actively deploying a spectrometer~\cite{smith2007probabilistic}. The problem was framed as a Partially Observable Markov Decision Process~(POMDP). While this is a principled approach to tackle sequential decision making problems in general, the POMDP framework suffers from scalability issues. Even with heuristics, the approach struggled to provide solutions in a reasonable time for environment sizes larger than $10\times 10$ grids. 

Thompson et al. applied Gaussian processes~(GPs) to create maps of geological phenomena in the environment by adaptively taking measurements with a spectrometer~\cite{thompson2008intelligent,thompson2011autonomous}. The key idea was to first learn a mapping between the spatial coordinates, on-board sensors and the geological phenomena using GPs, and then apply this GP online to determine where to take informative measurements to learn about the geological phenomena in a resource efficient manner. This is conceptually similar to the scientific information gathering problem we discuss in this paper. 

While GPs can effectively capture spatial relationships in a probabilistic manner, there are fundamental shortcomings to the GP framework which make Thompson's approach not directly applicable to our problem. Firstly, in this paper we are interested in encoding domain knowledge into the robot's decision making process. GPs can encode knowledge by imposing priors on the co-variance parameters, transforming the training data, and biasing the mean function~\cite{azmanincorporating}. However, here we are primarily exploring causal knowledge which GPs cannnot directly encode without significant computational overhead~\cite{lawrence2004gaussian}. Secondly, in general, the computational complexity of inference in GPs grows cubically with the number of observations made~\cite{rasmussen2006gaussian}. While there are ways to reduce this complexity, it remains computationally challenging to perform long horizon planning without making stationarity assumptions about the co-variance function. Scalability issues will be further compounded when the robot has to jointly plan with multiple sensing modalities especially when there are sensor correlations present~\cite{das2013hierarchical}. Our approach in contrast does not grow in complexity with the number of observations made, grows linearly with planning horizon, is anytime, and can be extended to an arbitrary number of sensing modalities without algorithmic modifications (discussed in Sec.~\ref{sec:approxreasoning}). 

\subsubsection{NASA OASIS}
OASIS contains two major subsystems: AEGIS~\cite{estlin2012aegis} and CASPER~\cite{knight2001casper}. The purpose of AEGIS is to automatically detect scientifically interesting features in the environment encountered during traversals through various computer vision techniques, and to take follow up measurements by pointing remote sensing instruments such as spectrometers. It has been deployed successfully on both Opportunity and Curiosity Mars rovers, and improved the science return by reducing the need to wait between command cycles for scientists on Earth to manually analyze rover imagery for interesting targets~\cite{francis2017aegis}.

AEGIS requires scientists to either specify interesting features a priori through a weighted function (for example, large rocks with high albedo may be interesting at a particular point in the mission), show examples of what is scientifically interesting by annotating a subset of images, or automatically detect anomalous observations. Like our approach, AEGIS biases data collection towards scientifically interesting areas. However, it does this indirectly while we directly optimizes for the science variable of interest in a principled manner which allows us to gather information in a more resource efficient manner. Our approach also handles noisy data robustly, and can incorporate arbitrary number of sensing modalities without adding additional heuristics.

CASPER is a planning and scheduling system which is currently been used on both Mars rovers and Earth Observing-1, a satellite which monitors natural events such as volcanic eruptions and floods~\cite{chien2004eo}. However, CASPER is a high level task planning system and fundamentally different to the planning we discuss in this paper. Given a set of goals and tasks, CASPER orders and prioritizes them in a way which is feasible under the resource and temporal constraints of the mission. In this paper, we are interested in determining a sequence of informative sensing actions, which requires reasoning about unobserved parts of the environment, sensor noise and the robot's current belief of the world, which is beyond the scope of task scheduling systems. 

\subsubsection{ESA ExoMars Rover Project}
The ExoMars Rover mission aims to send a rover to study signs of extinct life on Mars. There are several approaches that have been proposed to achieve science autonomy. The CREST Autonomous Robot Scientist project~(ASRP) has a similar architecture to OASIS-- a computer vision module to extract geological features and determine points of scientific interest, and an AI task planner similar to CASPER~\cite{woods2009autonomous}. ASRP developed techniques to extract higher level features such as salt deposits and cross-bedding from images, but utilized a weighted function to determine scientific value, and therefore has similar disadvantages to AEGIS. 

More recently, KSTIS, a fuzzy, knowledge based expert system was developed to emulate how a planetary geologist would assess a scene~\cite{barnes2009autonomous}. Although fuzzy systems provide more sophisticated identification of scientifically valuable targets than weighted functions, they require careful synthesis of rules to be effective, which is nontrivial in remote and previously unknown environments. Prior scientific knowledge, especially when it has inaccuracies, can be encoded more intuitively in our approach and errors in initialization can be automatically fine tuned as additional data is collected during the mission. Further, it is unclear how fuzzy systems will be able to reason about unobserved areas in the environment, and actively plan actions to reduce uncertainty. 

\subsubsection{Other work} 
Gallant et al. used a BN to classify rocks and assign benefit scores based on the current scientific goals of the mission~\cite{gallant2013rover}. The benefit scores were then fed into a cost function to determine the best action to take. However, their approach does not reason about unobserved parts of the environment and does not consider the problem of selecting which sensor to use. Pedersen et al. also used BNs to successfully classify rocks but did not feed back the probabilities to actively plan paths~\cite{pedersen2001autonomous}. Post et al. used BNs to create an obstacle map while integrating any sensor uncertainties that are present~\cite{post2016planetary}. A path was then planned to achieve a goal position while minimizing the probability of collisions. The work, however, does not attempt to model scientific knowledge, especially the spatial relationships often present in natural environments. More recently, Candela et al. explored a similar geological classification scenario to our Mars Lab experiments in Sec.~\ref{sec:autoscience} ~\cite{candelaIROS}. Like our approach, Candela also used BNs for classification and applied existing information gain path planning algorithms for efficient sampling. However, our approach is generalizable to arbitrary probabilistic relationships between variables, and enables planning with multiple sensing modalities and long horizons in a scalable and anytime manner.

\subsection{Knowledge Representation and Reasoning in AI}
The ideas of knowledge representation~(KR) and autonomous reasoning tools have been around since the beginning of AI research but first grew popular in the 1980s when foundational research in rules, frames and logical reasoning was conducted~\cite{brachman1985overview}. The developed frameworks achieved success in constrained problem domains such as medical diagnosis~\cite{kononenko1993inductive}; extending these approaches to more general tasks such as commonsense reasoning remains an active area of research~\cite{sowa2014principles}. 

Mainstream robotics, however, has had less success with KR frameworks. The key reason behind this is the uncertain and partially observable nature of real world environments. 
To successfully utilize KR frameworks in the real world, robots need to have two capabilities: (1) handle inconsistencies in data and the knowledge base and (2) update and refine the knowledge base over time as the robot acquires new information. This leads to an explosion of the robot's state space, and reasoning about the knowledge base to generate plans becomes computationally intractable and difficult to scale to large environments where information gathering robots typically operate~\cite{zhang2015mixed,hanheide2011exploiting}. 

We loosely base our approach to scientific information gathering on KR systems. We represent scientific knowledge as a BN and do approximate reasoning using MCTS techniques to plan future actions. BNs are limited in expressiveness as compared to other KR languages in AI literature, but suitable for modeling many aspects of scientific knowledge, in particular causal knowledge, the main type of knowledge we explore in this paper~\cite{hunter2010survey}. Being Bayesian in nature, they also handle uncertainty robustly. Due to these properties, many authors have employed BNs to model domain knowledge, particularly in diagnosis and expert systems applications~\cite{liedloff2013integrating,castillo2012expert}. Applying these networks to robotic decision making problems in unstructured environments is, however, less studied. Most authors have limited their use to classification and have not closed the loop around path planning~\cite{apostolopoulos2001robotic,sharif2015autonomous}. There is also extensive research in approximate inference techniques for BNs which we can directly exploit for better scalability and real time performance~\cite{beal2003variational,murphy1999loopy}. We essentially trade off potential expressiveness for greater robustness to uncertainty, and the ability to plan long horizons quickly. 

\subsection{Informative Path Planning}
\label{subsec:infoplanning}
Exploration and sensor planning to gain information about the world can be seen as an informative path planning problem. However, multi-modal scientific information gathering is richer in scope than traditional literature as it requires optimizing information gain with respect to a variable which cannot directly be measured, involves planning with multiple on-board sensing modalities which measure different variables, and is often conducted in partially observable environments. 

Greedy approaches are often effective approaches for information gathering and offer performance guarantees when the problem is submodular~\cite{krause2007near}. Unfortunately, these performance guarantees are lost when path dependent costs are present. Branch and bound techniques which prune suboptimal branches early in the tree search have shown promise~\cite{binney2012branch,bestprobabilistic} but efficiently calculating tight bounds in problems with unknown environments and multiple sensing modalities is nontrivial. Approaches that involve initially unknown environments often utilize GPs and exploit the monotone sub-modular nature of the mutual information or variance reduction function to avoid exhaustive search~\cite{hollinger2014sampling}. GPs however have the disadvantages mentioned in Sec.~\ref{subsec:lifeatacama}. There are also application specific approaches in literature but they either do not generalize to unknown environments or cannot plan for multiple sensing modalities without additional heuristics~\cite{jawaid2015informative}. 

In field applications of information gathering, several approaches have been proposed. Thompson et al. used a greedy algorithm to design maximally informative trajectories for constructing spatial maps of multi-spectral data \cite{thompson2008intelligent}. Wettergreen et al. extended this in \cite{wettergreen2014science} to design trajectories that explore regions of orbital maps that cannot be explained with previous observations, actively solving the spectral unmixing problem. Girdhar et al. used a database of observations to detect anomalous data. Similar to our approach, a generative Bayesian model was updated online by directing the robot towards these anomalies~\cite{girdhar2016modeling}. However these approaches used short planning horizons and do not make decisions about using secondary sensors to gain information. 

Tabib et al. explored a search and rescue application where their robot planned trajectories that maximize the information gained by two different sensors which measure the geometry and temperature of the environment~\cite{tabib2016efficient}. It is assumed that the instruments are constantly collecting data, instead of being actively switched on and of. This assumption simplifies the planning problem but is not practical in situations where deploying sensors is expensive and requires the robot to be stationary (i.e. a soil sampling mechanism). 

MCTS methods have recently been applied to the robotic informative path planning domain~\cite{best2016decentralised,patten2017monte}. They work for any general objective function and do not require bounds. They are also anytime and hence suitable for online planning~\cite{browne2012survey}. In this paper, we show how MCTS can be adapted to reason about scientific knowledge, plan with multiple sensing modalities, and remain robust to any uncertainty propagated into the system as a result of partial observability and noisy data. 

\begin{figure*}[ht]
\centering
    \subfloat[b][]{
    \centering
    	\includegraphics[trim = {7cm 4.5cm 7cm 5cm},clip,width=0.6\textwidth]{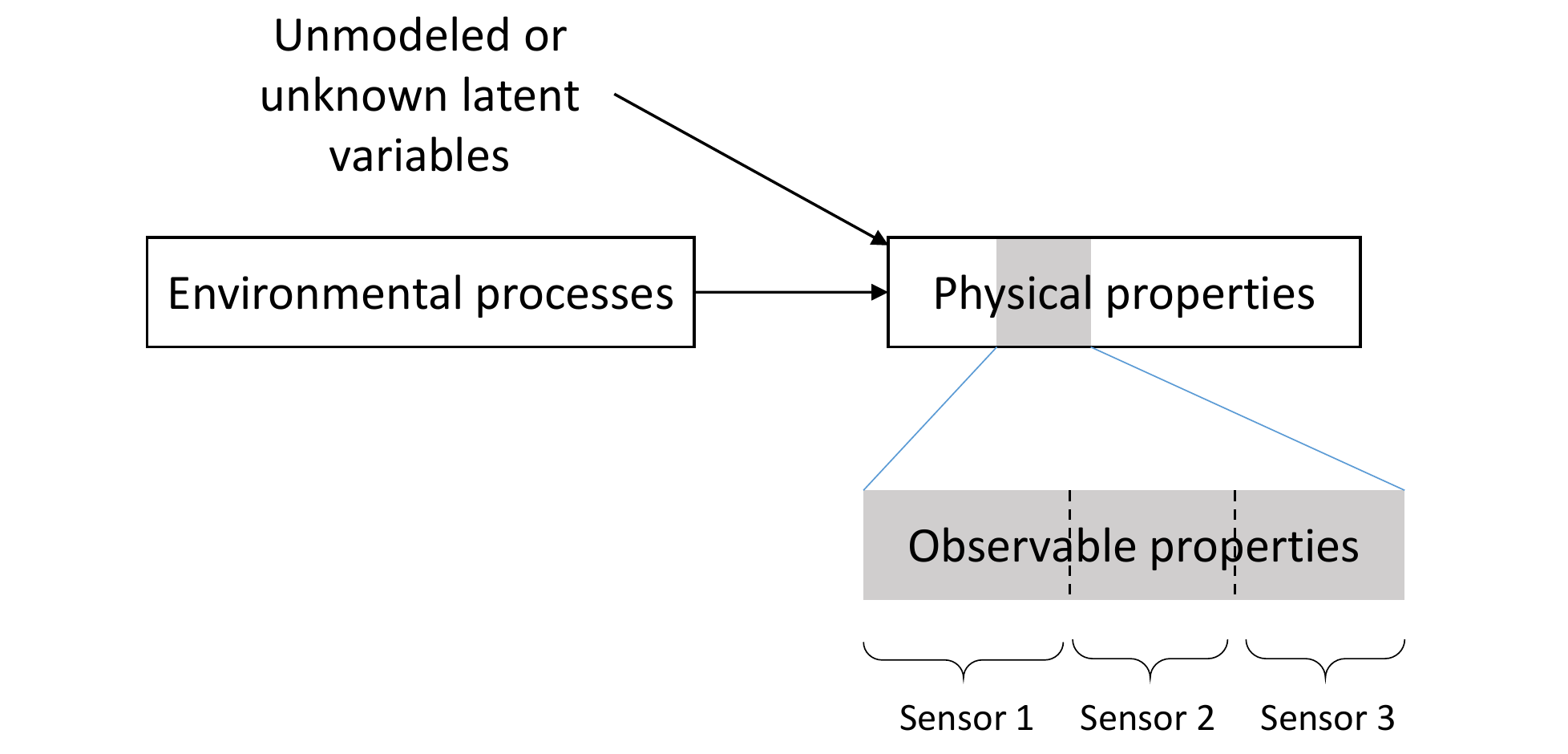} \label{fig:science_knowledge}}
    ~ 
    \hfill
    \subfloat[b][]{
    \centering
        \includegraphics[trim = {10cm 5cm 10cm 5cm},clip,width=0.4\textwidth]{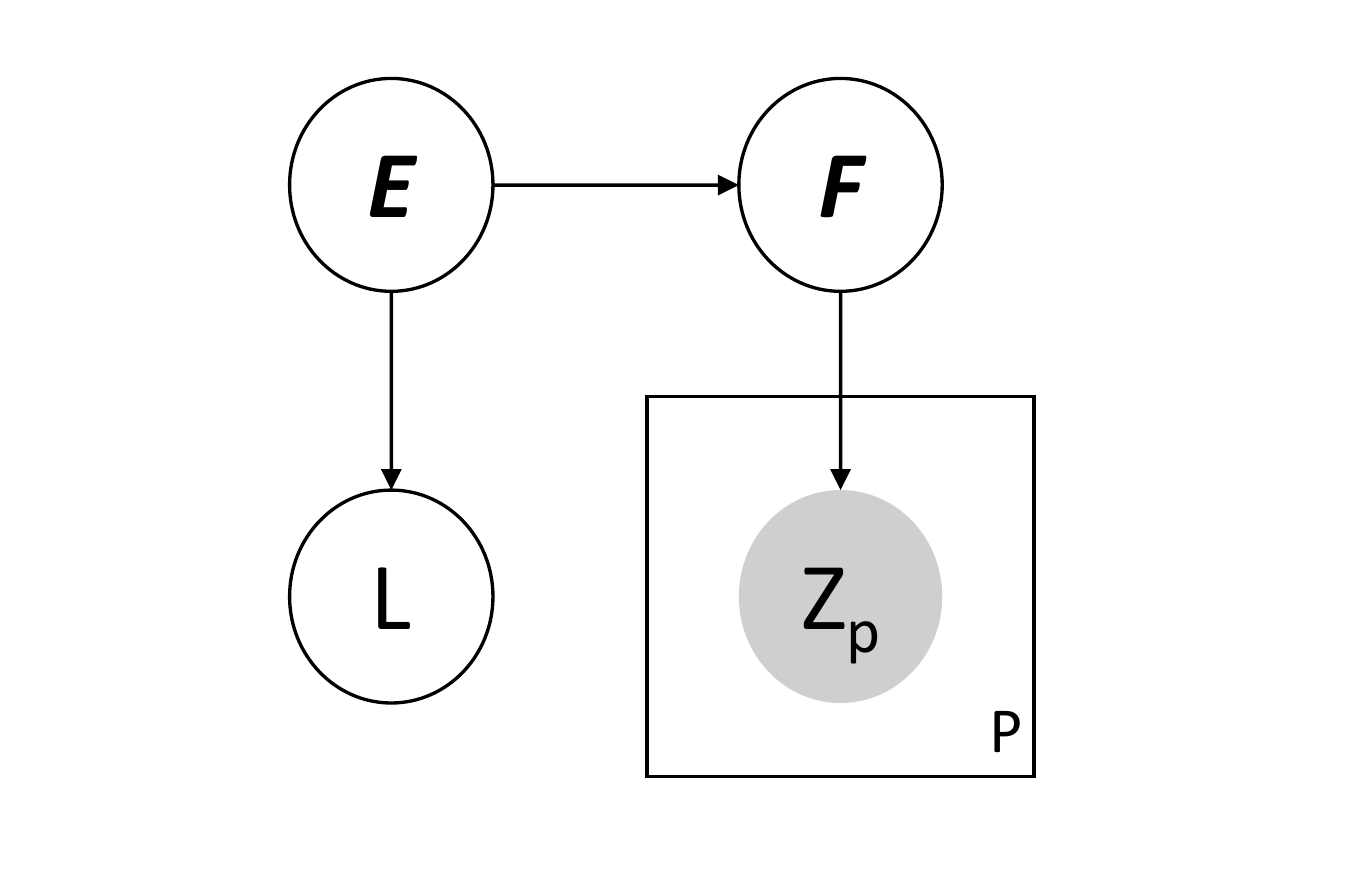}\label{fig:science_knowledgeBN}}
\caption{(a) A typical structure of scientific knowledge in a geological context. (b) The generalized BN approximation we use in this paper. Physical properties are represented by feature vector $F$. A subset of the feature space can be observed with the sensing modalities on-board. An observation made by sensor $p$ is denoted $Z_p$ and there are $P$ such sensors available. $L$ is the scientific variable of interest which is characterized by the underlying environmental processes $E$. The BN maintains the strong causal relationships in the environment through its directed structure} 
\end{figure*}

\section{Multi-Modal Scientific Informative Gathering}\label{sec:problem_def}

In this section we formally define our planning problem and outline a generic solution structure. Similar to informative path planning, multi-modal scientific information gathering is, at its core, a sequential decision making problem. The robot is required to select a sequence of sensing actions $a_{seq}$ out of the space of all possible actions $A$, which maximize the information gained. However, one important distinction is that we are interested in gaining information about some latent variable of scientific interest $L$, which cannot necessarily be directly observed using onboard sensors.

A second defining characteristic is that the robot has access to multiple sensing modalities. Each sensing action is represented by a tuple that consists of a motion primitive (chosen from a finite discrete set defined a-priori based on the robot's motion model) and the type of sensor used. There is also an energy or time cost associated with each action, and the robot is constrained to a total sensing budget $S$ driven by mission requirements. The optimization objective is then given by

\begin{equation}
\begin{split}
&a^*_{seq} = \operatorname*{arg\,max}_{a_{seq} \in A} EI(a_{seq}) \\
&\textbf{s.t.} \sum^{|a_{seq}|}_{i}{\textnormal{cost($a_i$)} = S}\,,
\end{split}
\label{eq:optimisation}
\end{equation}
where $EI$ is the expected informativeness of an action sequence. This is calculated by marginalizing out all possible observations $Z_{seq}$ that can result from the sensing sequence
\begin{equation}
EI(a_{seq}) = \sum_{Z_{seq}}{I(Z_{seq})P(Z_{seq}|a_{seq})}\,,
\label{eq:expected}
\end{equation}
where $I$ is a function which measures informativeness of sensing actions. We use traditional Shannon information gain given by
\begin{equation}
I(Z_{seq}) = H(L) - H(L|Z_{seq})\,,
\label{eq:infodef}
\end{equation}
where $H$ is the Shannon entropy. Lastly, in this paper we consider discrete world environments where the robot is required to estimate the state of $L$ in each of the $N$ cells. Eq.~\ref{eq:infodef} can thus be further decomposed into
\begin{equation}
I(Z_{seq}) = \sum_{n=1}^{N}{H(L_n) - H(L_n|Z_{seq})}\,.
\label{eq:infodiscrete}
\end{equation}


To solve this optimization, two key capabilities are required. The robot firstly needs to evaluate the conditional entropy term $H(L|Z_{seq})$ for which some mapping from a given observation sequence to the latent variable $L$ is required. This mapping is provided by the scientific knowledge representation framework detailed in Sec.~\ref{sec:knowledgerep}. The robot then needs to conduct a nonmyopic (long horizon) search over the heterogeneous sensing action space to determine informative action sequences. Our planning approach is detailed in Sec.~\ref{sec:approxreasoning}. 

\section{Representing Scientific Knowledge}
\label{sec:knowledgerep}
In this section, we describe a method to encode scientific prior knowledge using a Bayesian network for the purpose of evaluating the conditional entropy term $H(L|Z_{seq})$, a key requirement in determining informative sensing sequences. We show how a tree-structured BN effectively captures relevant aspects of scientific knowledge while keeping inference quick.


An example representation of prior scientific knowledge in a geological context is shown in Fig.~\ref{fig:science_knowledge}. There is a wide range of physical properties that describe the environment, out of which a subset can be observed depending on the sensing modalities available to the robot. The interaction of environmental processes characterize the distributions of these physical properties. For example, extended periods of heavy wind over a geological region affect the amount of erosion visible on rock surfaces. There are also likely to be latent variables which cannot be modeled easily or simply unknown a priori. Since the goal of science exploration missions is to often to acquire new knowledge, there are also likely to be inaccuracies and gaps in prior scientific knowledge. 

A key advantage of a BN representation here is that relationships between environmental and other mission variables are represented as conditional probabilities. Thus, inference remains robust to modeling uncertainties and noisy observations. Another advantage is that BNs can naturally capture the causal structure between variables through the directed edges. We propose the network structure shown in Fig.~\ref{fig:science_knowledgeBN}. 

Each of the $P$ onboard sensors can make some observation $Z_p$ which reveals information about the feature space vector $F$. The observations are propagated through the network to make inferences about any environmental processes $E$ and the latent variable of mission interest $L$. Quantitative knowledge can be fed into the network by initializing the conditional probability parameters as well as the prior probability distributions of parent nodes. Depending on the application, the science variable of interest could be an environmental process, a physical property, or something more abstract such as evidence for life. This framework allows the robot to work with any such mission requirement and even switch between science variables during the mission without any algorithmic modifications. In this paper, we assume all variables modeled in the BN are categorical variables, but with some minor modifications, our approach can also be extended to continuous variables.

Our framework allows for arbitrary BNs or even general factor models to be used. However, to make inference and subsequent planning efficient, there are several design considerations to be made. Our proposed network has a bottom-up tree-like structure where sensor observations are made at leaf nodes of the tree and the latent variable of interest is near the root of the tree. This structure achieves quick inference times by enabling the application of the message-passing technique~\cite{yedidia2000generalized}, which allows exact inference to be done in $2$ passes through the network. The tree-like structure allows recursive updating of nodes, which removes the need to track observation history, and also avoids cycles in the graph that have been known to exact inference difficult. 

In field environments, the number of environmental variables as well as vector sizes could be very large. To keep belief updates tractable, it is important to collapse the network nodes until only the key influential nodes are left. Furthermore, since the BN is intended as an approximation, the network structure does not have to exactly match the true causal relationships in the environment. The network can be transformed by reversing edges and marginalizing out nodes to achieve faster inference. The trade-off between model accuracy and computational efficiency remains an application dependent choice.  

\section{Approximating Planning}
\label{sec:approxreasoning}
In this section, we highlight the computational challenges in solving Eq.~\ref{eq:optimisation} and then propose sampling-based solutions to reason about BNs and determine informative action sequences. 

\subsection{Complexity analysis}
Combining equations \ref{eq:optimisation}, \ref{eq:expected} and \ref{eq:infodef} and dropping the constant terms, the optimization can be simplified to
\begin{equation}
\begin{split}
a^*_{seq} &= \operatorname*{arg\,max}_{a_{seq} \in A} \sum_{Z_{seq}}{-H(L|Z_{seq})P(Z_{seq}|a_{seq})} \\
&\textbf{s.t.} \sum^{|a_{seq}|}_{i}{\textnormal{cost($a_i$)} = S}\,.
\end{split}
\label{eq:newoptimisation}
\end{equation}

There are three main components involved in deducing the optimal sensing sequence: (1) calculating the conditional entropy of the latent variable of interest given some observation sequence, $H(L|Z_{seq})$, (2) repeating this calculation and summing over all possible observations that can result from an action sequence, and (3) iterating the process over the action space to determine the best sensing actions to take. We discuss each of these components, illustrate the challenges involved in optimization, and introduce our sampling based approaches. 

\textbf{Calculating conditional entropy:} Using the definition of Shannon entropy, the $H(L|Z_{seq})$ term can be expanded into
\begin{equation}
H(L|Z_{seq}) = -\sum_{L}{P(L|Z_{seq})\ \log{P(L|Z_{seq})}}\,.
\label{eq:entdef}
\end{equation}

By exploiting the structure of the scientific knowledge BN, the $P(L|Z_{seq})$ term can be factorized further. We use the BN in Fig.~\ref{fig:science_knowledgeBN} as an example to give insight into the mathematics and computational complexity
\begin{equation}
\begin{split}
P(L|Z_{seq}) &= \frac{P(L,Z_{seq})}{P(Z_{seq})} \\
&= \eta \sum_{E, F}{P(L,Z_{seq},E,F)} \\
&= \eta \sum_{E}{P(E)P(L|E)\sum_{F}{P(F|E)\prod_{i}^{|seq|}{P(Z_i|F)}}}\,,
\end{split}
\label{eq:condent}
\end{equation}

where $\eta$ is a normalization constant. Solving Eq. \ref{eq:condent} exactly requires summing over the space of values that $F$ and $E$ can take. If the feature space $F$ is a vector of $N$ features $[f_1, f_2, ...f_N]$, there will be $|f|^N$ possible instantiations of the feature space where $|f|$ is the number of categories each feature can take. In field applications, $|F|$ may grow very large, especially if visual or hyper-spectral sensors are involved. Fortunately, approximate inference in Bayesian networks is a well-studied problem and we can directly apply a number of approaches to approximate Eq. \ref{eq:condent} to satisfy any accuracy and computational time requirements of the mission~\cite{beal2003variational,murphy1999loopy}. For tree-like BN structures such as the one proposed in Fig.~\ref{fig:science_knowledgeBN}, the message passing technique can efficiently propagate belief updates through the network~\cite{yedidia2000generalized}. The computational time required to calculate the conditional entropy for a single observation sequence is denoted by $T_L$.

\textbf{Predicting and summing over observations: }The next component is evaluating the conditional entropy term over all possible observations that can result from an action sequence. The $P(Z_{seq}|a_{seq})$ term is calculated by first convolving the fields-of-view of sensors with the robot's path to determine what areas of the environment will be seen during the sensing action, and then applying any sensor noise; it is effectively a sensor model term. However each sensing action can potentially produce high dimensional or even continuous observations, and the space of possible observation sequences grows exponentially with the length of the sensing sequence. Performing this calculation exactly is therefore not practical. 

\textbf{Iterating over the action space: } As mentioned earlier, the action space $A$ of the robot in a single decision step is of size $|M|\times|S|$ where $|M|$ and $|S|$ are the number of motion primitives and number of sensors that can be used respectively. For a planning horizon $H$, the total time complexity of an exact solution to Eq.~\ref{eq:newoptimisation} is $O((T_L.|Z|.|A|)^H)$ which quickly becomes intractable to solve in the presence of large environments, long mission durations and large observations spaces. We present two approximate approaches: a greedy solution and an MCTS-based online planning algorithm that updates and adapts plans as observations are taken. 

\begin{figure*}[ht]
  \includegraphics[trim={4cm 5.3cm 0cm 1cm},clip,width=0.9\textwidth]{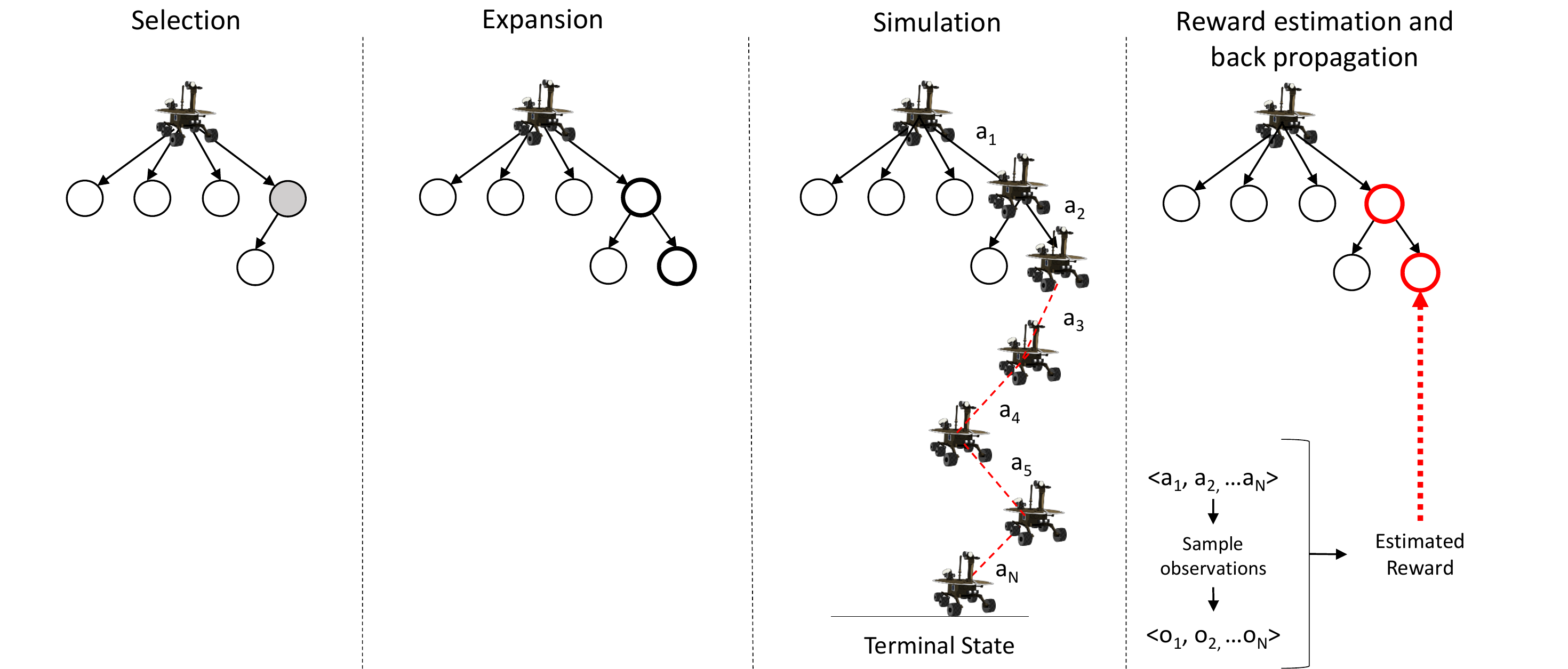}
\centering
\caption{The four stages of MCTS applied to the multi-modal scientific information gathering. The stages are repeated until some computational budget has expired, at which point the child of the root node with the highest average reward is returned as the best sensing action to execute.}
\label{fig:mcts}       
\end{figure*}

\subsection{Greedy Optimization}
\label{sec::greedy}
The greedy version of the optimization is given by Eq.~\ref{eq:greedyopt}. We select the sensing action with the highest information gain to cost ratio 
\begin{equation}
\begin{split}
a^*_{next} &= \operatorname*{arg\,max}_{a \in A} \sum_{Z_{a}}{U(Z_a) P(Z_a| a)} \\
&\textbf{s.t.} \textnormal{ cost($a^*_{next}$)} \leq \text{Leftover Budget}\,,
\end{split}
\label{eq:greedyopt}
\end{equation}
Where:
\begin{equation}
U(Z_a) = \frac{H(L) - H(L|Z_a)}{Cost(a)}\,.
\end{equation}

As mentioned earlier, the observation space can potentially be very large and evaluating Eq.~\ref{eq:greedyopt} exactly in not practical. Monte Carlo sampling is therefore used to offset the high dimensionality of the observation space. For each sensing action, forward kinematics are first applied to determine which cells of the environment will be seen. $N_S$ samples are then drawn from the current belief of the observation space in the corresponding cells. The utility of each sampled observation is calculated by simulating a belief space update and calculating the information gained. As the number of samples increase, the average information gained from the sampled observations converges to the true expected utility of actions:
 
\begin{equation}
\begin{split}
E[U(a)] &= \sum_{Z_a}{U(Z_a).P(Z_a|a)} \\ 
		&\approx \frac{1}{N_s}\sum_{i}^{N_S}{U(Z_i)}\,.
\end{split}
\label{eq:utilitysample}
\end{equation}

The sensing action with the maximum expected utility is executed and the robot takes an observation using one of its sensors. If the knowledge BN is appropriately structured, the belief on the latent variable of interest can be updated recursively and removes the need to track past observations. Pseudocode is given in Alg.~\ref{alg:single_step_planner}. 

\begin{algorithm}[t]
\caption{Pseudocode for a single step planner}\label{alg:single_step_planner}
\begin{algorithmic}[1]
\State \textbf{Input:} SensingBudget $S$, BeliefSpace $Bel$, DomainKnowledge BN $K$, RemainingBudget $R$
\Function{Main}{} 
	\State $R \gets S$
	\While{$R > 0$}
		\State $robotPose \gets getLocalisation()$
        \State $a_{opt} \gets greedyPlanner(robotPose, R, Bel, K)$
        \State $Z\gets takeObservation(a_{opt})$
		\State $Bel\gets updateBeliefSpace(Z, Bel, K)$
        \State $R \gets R - cost(a_{opt})$
	\EndWhile
\EndFunction
\State
\Function{greedyPlanner}{$robotPose, R, Bel, K$}
\State $a\gets generateSamples(A) $ \Comment{Sample action space}
\For{i=1:size(a)} \Comment{Iterate through each action}
\State $P(Z_s)\gets sampleObservations(a_i, B)$ 
\State $E[U(a_i)]\gets calculateExpectedUtility(Z_s, B)$
\EndFor{}
\State \textbf{end}
\State $a_{opt} =  \operatorname*{arg\,max}_{a \in A} EU(a)$
\State
\Return $a_{opt}$
\EndFunction
\end{algorithmic}
\end{algorithm}

\subsection{MCTS Non-myopic Optimization}
\label{sec::MCTS}
In information gathering missions, the robot acquires observations after executing every sensing action. At the end of planning time, the robot only needs to commit to the first action of the sensing plan and has the freedom to adapt the sensing plan after the observation is taken. We can therefore treat this as a sequential decision making problem. To approximate the solution, we propose the use of the MCTS algorithm. The algorithm adapted to our problem is presented in Alg.~\ref{alg:mcts}. 

MCTS involves cycling through four stages: node selection, expansion, simulation and back-propagation. The key idea is to first select promising leaf nodes based on a tree policy. The selected node is expanded and a terminal reward is estimated by conducting simulations or ``rollouts" in the decision space. The reward of the rollout is calculated and back-propagated up the tree. The process is repeated until some computational budget is reached. At the end of the search, the child of the root node with the highest average reward is selected as the next best action.

\begin{algorithm}[t]
\caption{MCTS Science Autonomy Planner}\label{alg:mcts}
\begin{algorithmic}[1]
\State \textbf{Input:} SensingBudget $S$, BeliefSpace $Bel$, DomainKnowledge BN $K$, RemainingBudget $R$
\Function{Main}{} 
	\State $R \gets S$
	\While{$R > 0$}
		\State $robotPose \gets getLocalisation()$
        \State $a_{opt} \gets planner(robotPose, R, Bel, K)$
        \State $Z\gets takeObservation(a_{opt})$
		\State $Bel\gets updateBeliefSpace(Z, Bel, K)$
        \State $R \gets R - cost(a_{opt})$
	\EndWhile
\EndFunction
\State
\Function{planner}{$robotPose, R, Bel, K$}
    \State $T \gets initializeTree(robotPose, R)$
	\State $currentNode \gets T.rootNode$
	\While{within computational budget}
		\State $currentNode \gets treePolicy(T)$
		\State $sequence\gets rolloutPolicy(currentNode, R)$
        \State $reward \gets getReward(sequence, Bel, K)$
        \State $T \gets updateTree(T, reward)$
	\EndWhile
    \State
    \Return $bestChild(T)$
\EndFunction
\State
\Function{rolloutPolicy}{$currentNode, R$}
	\State $sequence \gets currentNode$
    \While{$R > 0$}
    	\State $nextNode \gets defaultPolicy(currentNode)$
        \State $currentNode \gets nextNode$
        \State $sequence \gets sequence + currentNode$
        \State $R \gets currentNode.R$
    \EndWhile
    \State
    \Return $sequence$
\EndFunction
\State
\Function{getReward}{$sequence, B, K$}
	\State $reward \gets 0$
    \For{$i=1:length(sequence)$}
    	\State $currentAction \gets sequence(i)$
        \State $Z = sampleObs(currentAction, Bel, K)$
        \State $Bel_{new} = updateBelief(Z, Bel, K)$
        \State $infoGain = calcInfoGain(Bel_{new}, Bel)$
        \State $reward \gets reward + infoGain$
        \State $Bel \gets Bel_{new}$
    \EndFor
    \State
    \Return $reward$

\EndFunction

\end{algorithmic}
\end{algorithm}

We formulate the MCTS such that each node in the tree is a potential sensing action. It is a tuple consisting of the robot's x and y positions, the orientation, the type of sensor used and the remaining sensing budget. Each node also stores the average reward $\bar{R_i}$ of all the simulations that have passed through it and the number of times it has been visited $n_i$. The children of the node are determined by the robot's action space and the remaining budget. We now describe each stage of the MCTS in detail and show how it has been adapted for our problem. 

\textbf{Selection:} The first stage of MCTS is to use a tree policy to select which leaf nodes to expand. We want to expand leaf nodes which are expected to have a good terminal reward but at the same time evaluate alternative nodes sufficiently to minimize chances of converging to local minima. The Upper Confidence Tree~(UCT) policy based on the optimism in the face of uncertainty paradigm is known to be a good solution to balance the exploration/exploitation trade-off present here~\cite{kocsis2006bandit}. UCT begins at the root node and iteratively selects leaf nodes with the highest Upper Confidence Bound~(UCB) until a node with unexpanded children is reached. The UCB score for node $i$ is defined by
\begin{equation}
UCB_i = \bar{R_i} + C_p\sqrt\frac{2\log N}{n_i}\,.
\label{eq:UCB}
\end{equation}

The first term is the ``exploitation" component of UCB, where $\bar{R_i}$ is the average reward of all rollouts that have passed through $node_i$. We define the reward function below. The second term in the equation is the ``exploration" component where $N$ is the number of times the parent of the node has been evaluated and $n_i$ is the number of times node $i$ has been evaluated. $C_p$ is a constant that balances exploration and exploitation. It is usually selected such that it is on a similar scale as the typical rewards in the problem. We found empirically that a value of $0.1$ gave good results in both simulations and hardware experiments. 

\textbf{Expansion:} From the leaf node selected by the UCT policy, an unexpanded child node is randomly selected based on the action space and added to the tree. 

\textbf{Simulation:} The aim of the simulation stage is to determine the terminal reward associated with this newly expanded child node by executing some rollout policy. Here we use a random action selection policy from the selected node until the sensing budget is exhausted. A random policy was used because it requires minimal computational overhead to calculate and ensures the decision space is uniformly explored. However a large number of rollouts are often required to accurately estimate rewards. Using problem specific rollout policies has been shown to significantly improve tree convergence~\cite{gelly2012grand} but we leave this as an interesting avenue for future work. 

The ideal reward function to evaluate a rollout would be the expected information gain defined earlier in Eq.~\ref{eq:expected} as this is the function we are directly interested in optimizing. Evaluating this function exactly however requires summing over all possible observations that can result from the rollout sequence, which from previous analysis in Sec.~\ref{sec:approxreasoning} was deemed to be computationally intractable to evaluate. 

We define the reward of a rollout as $\frac{I_r}{H_{init}}$ where $I_r$ is the information gain during rollout $r$ and $H_{init}$ is the joint entropy of the $L$ variables at the current state of the mission. This division constrains the average reward to between $0$ and $1$: a requirement for UCB convergence guarantees to hold. We approximate the information gain of the rollout by sampling. We begin at the first node of the rollout. An observation is sampled from the belief space corresponding to the sensing action used. The belief space is updated and passed onto the next node. The process is iterated until the last node of the rollout sequence is reached and the total information gain is determined by subtracting the entropies of the initial and final belief space. 


\textbf{Back-propagation:} Lastly the reward received by the rollout is back-propagated up the tree and the average reward and number of evaluations for each node involved in the rollout is updated. 

The four stages are repeated until the computational budget for the robot has expired, at which point the root node's child with the highest average reward is selected as the next best action. Given enough samples and an appropriate value for the exploration parameter $C_p$ in Eq.~\ref{eq:UCB} it can be shown that the tree will converge to the optimal action sequence~\cite{kocsis2006bandit}. This formulation gives us a principled approach to incorporate an arbitrary number of sensors in planning as each sensing action is simply a branch in the tree and does not require the creation of additional heuristics to handle. Long horizons and uncertainty are also handled robustly in an anytime manner. 

The total time complexity of the MCTS algorithm is $O(N_S.T_L.H)$ where $N_S$ is the number of iterations the MCTS is run, a significant improvement over the brute force search complexity of $O((T_L.|Z|.|A|)^H)$. It is important to note, however, that large action spaces mean that MCTS will take more iterations to explore the deeper nodes in the search tree while problems with large observation spaces require more iterations to accurately estimate rewards. The trade-off between optimality and computational time remains an application-dependent choice. 

\section{Active Information Gathering for Mars Exploration}
\label{sec:autoscience}
In this section, we demonstrate how our approach can be implemented and applied in a Mars robotic exploration scenario, and quantify the performance benefits over passive information gathering approaches in both simulation and planetary rover experiments.   

\subsection{Scenario Overview}
\label{sec::marsexploration}
Deducing the geologic type of a location (i.e. riverbed, desert or volcanic regions) is fundamental to many science objectives of Mars exploration missions~\cite{grotzinger2012mars}. We therefore consider a problem where the robot is required to learn about the types of locations in the environment by taking measurements of geological features at informative locations through multiple sensing modalities. This section discusses the properties of our robot and the assumptions made about the world, and formally defines the planning problem that the robot is required to solve in this Mars exploration mission.

\textbf{Environment Setup: } We assume the simulated robot is a ground-based vehicle that moves around in a world discretized into cells. The environment is initially unknown to the robot. 

\textbf{Robot Properties:} While our approach works for an arbitrary number of sensors, we assume the rover here is equipped with two sensors. The first sensor is a camera that can detect rocks within its field-of-view and extract their visual features. The field-of-view may span multiple cells. The second sensor is an ultraviolet~(UV) light source that the robot can project onto the environment to reveal UV reflective minerals. The UV light source simulates what a spectrometer might do on a real Mars mission since it is energetically expensive to use and has a narrow sensing range, but gives more informative measurements than a camera. 

\textbf{Action Space:} A sensing action is characterized by a movement action and the type of sensor used.  There are two sensors the robot can choose from and we discretized the movement actions into 5 actions: moving forward one cell in the direction the robot is facing, or turn in increments of $-90, -45, 45$ and $90$ degrees. Thus the total action space is of size 10. We use a constant cost function where the cost of using the camera sensor is 1 unit and the cost of using the UV sensor is 8 units. The robot must use a sensor during each action.

\textbf{Optimization Objective:} Following on from Eq.~\ref{eq:optimisation}, the planning objective for this problem is:
\begin{equation}
\begin{split}
a^*_{seq} &= \operatorname*{arg\,max}_{a_{seq} \in A} \sum_{Z_{seq}}{P(Z_{seq}|a_{seq}).I(L|Z_{seq})} \\ 
&\textbf{s.t.} \sum^{|a_{seq}|}_{i}{\textnormal{cost($a_i$)} \leq S}\,.
\end{split}
\label{eq:autoscience}
\end{equation}

$I(L|Z_{seq})$ is the total information gain, calculated by summing the change in entropy in the type of location $L_n$ in each grid cell $n$:
\begin{equation}
I(L|Z_{seq}) = \sum_{n=1}^{N}{(H(L_n)-H(L_n|Z_{seq})}\,. 
\end{equation}

\subsection{Knowledge Representation}
This section discusses the BN we use to model scientific knowledge in this Mars exploration problem. The BN creates a probabilistic mapping between the observations made by the two sensors and the type of location in a cell, which can then be used by the robot to automatically identify and execute informative sensing actions. 

\begin{figure}[t]
\centering
\includegraphics[trim={5cm 3cm 8cm 6.5cm},clip,width=0.5\textwidth]{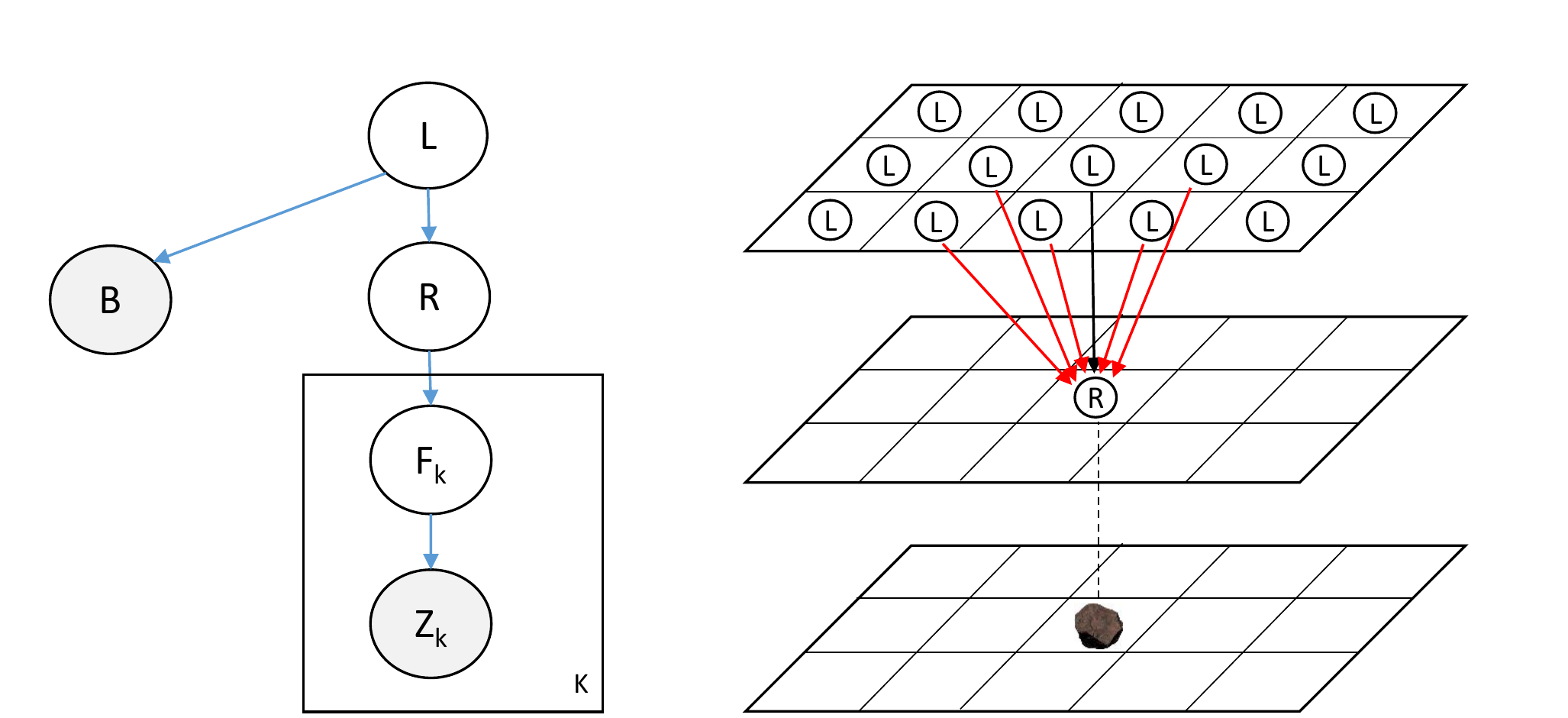}
\caption{Left: The structure of the Bayesian network used to represent geological knowledge. Right: Spatial relationships between adjacent cells}
\label{knowledgebn}
\end{figure}

The BN shown in Fig.~\ref{knowledgebn} is used. On Mars, rocks are the key sources of geological cues. The type or class of rocks in the environment is represented by variable $R$. Each rock exhibits $K$ visual features represented by variable $F$. The robot can observe these features through its camera sensor and this observation is denoted by $Z$. The variable $B$ is the UV reflective material that can be measured by the robot's UV sensor. Lastly, $L$ is the geologic type of the location the robot is currently in and this is the scientifically interesting latent variable we are interested in learning about. This causal dependency between the variables is reflected in the structure of the BN.

The proposed BN structure allows several sources of information to be integrated in the form of conditional probabilities. $P(Z|F)$ is the sensor model, $P(F|R)$ is the classifier likelihood while $P(B|L)$ and $P(R|L)$ are geological properties of the environment. All nodes in the network are discrete as geologists often look for features which do not necessarily have intuitive continuous measurements such as the presence of bedding or smoothness of a rock. Discretization also simplifies inference and allows the MCTS to conduct a larger number of forward simulations in a given time frame. 

In natural environments there are often strong spatial correlations present. There are several methods of encoding this relationship. A common approach is through a Markov random field where undirected edges are added between nodes in adjacent grid cells. Although, this is an expressive way to capture spatial dependencies, the inference problem becomes difficult as cycles are introduced in the graphical model. Another alternative is to add links between the $L$, $R$ and $B$ nodes of adjacent cells. When Bayesian updates occur, a spatial region in the robot's neighborhood is updated instead of just the nodes in the current cell. Nodes that are far from where the observation was taken are less affected. This decreasing influence is modeled by a Gaussian function. Fig.~\ref{knowledgebn} illustrates this spatial dependency. The grid resolution of each variable does not have to match and can be adapted based on the expected spatial variability.

The conditional probability parameters can either be specified directly through domain knowledge, learned from training data~\cite{heckerman1995learning} or even learned online by modeling them as Dirichlet distributions. In this scenario we assume that the maximum likelihood parameters are known a priori. Due to this BN's structure, the belief on the value of nodes can be updated recursively without keeping a history of observations.

\subsection{Simulation Experiments}

In these experiments, we aim to empirically demonstrate the performance of active approaches to scientific information gathering over passive approaches. As mentioned in Sec.~\ref{subsec:infoplanning}, there are several algorithms in the literature for informative path planning~\cite{hollinger2014sampling,binney2012branch}. However, these approaches are not suitable without significant algorithmic modifications for tackling situations in which the robot has to simultaneously plan informative paths which adhere to budget and goal constraints and decide \textbf{when} to activate \textbf{which} sensor in initially unknown environments. We therefore compare the performance of the following approaches:

\textbf{Random sampling:} the robot selects a random action within its action space at each time step which does not break the budget constraint. This serves as the baseline algorithm and is an example of passive information gathering. 

\textbf{Fixed sampling:} When one sensor is involved and a goal position constraint is given, a lawnmower pattern is popular as it provides uniform coverage. Designing sensing schedules for $N$ instruments with varying sensing costs is however non-trivial. Here we use a five-stage policy that involves the robot using the camera sensor in the forward direction, 90 degrees to the left, and 90 degrees to the right, using the UV sensor in the current cell and then moving one step forward. The stages are repeated until the robot's sensing budget is exhausted. This is an example of a traditional passive algorithm where scientists guide the robot through policies they believe to be informative. 

\textbf{Greedy:} The strategy introduced in Sec.~\ref{sec::greedy} which selects sensing actions with the highest immediate expected information gain to cost ratio. $20$ samples were used to estimate the utility of each action. The behavior is similar to a frontier-based strategy often used in exploration problems but allows incorporation of multiple sensors. This is an active algorithm. 

\textbf{MCTS:} The approach discussed in Sec.~\ref{sec::MCTS} which plans non-myopic sensing actions. It is an anytime algorithm and we use 100 iterations for comparison purposes. This relatively small number of iterations is reflective of the limited on-board computational resources of field robots.  

\begin{figure}[t]
\centering
\includegraphics[trim = {3cm 4cm 3cm 3.5cm},clip,width = 0.5\textwidth]{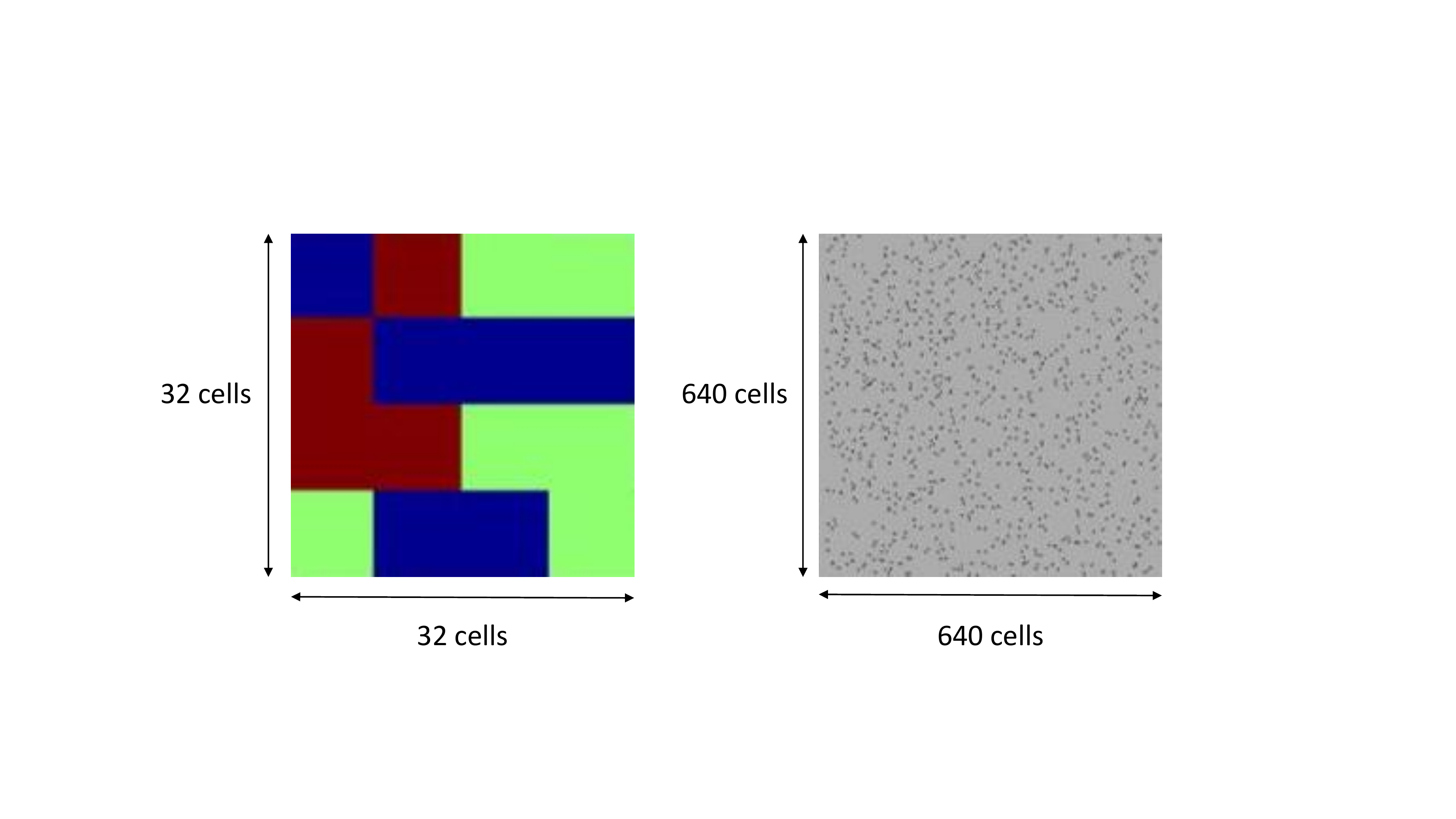}
\caption{Left: An example ground truth map for location type. Right: An instantiation of the rock grid}
\label{egmap}
\end{figure}

$32\times 32$ grid world environments were randomly generated in which the grid was further divided into 16 $8\times 8$ regions of homogeneous location types as shown in Fig.~\ref{egmap}. The robot occupies one cell in the grid at a time and can be oriented in one of 8 directions in $45\degree$ increments.

\begin{figure*}[t]
\centering
\includegraphics[width=\textwidth]{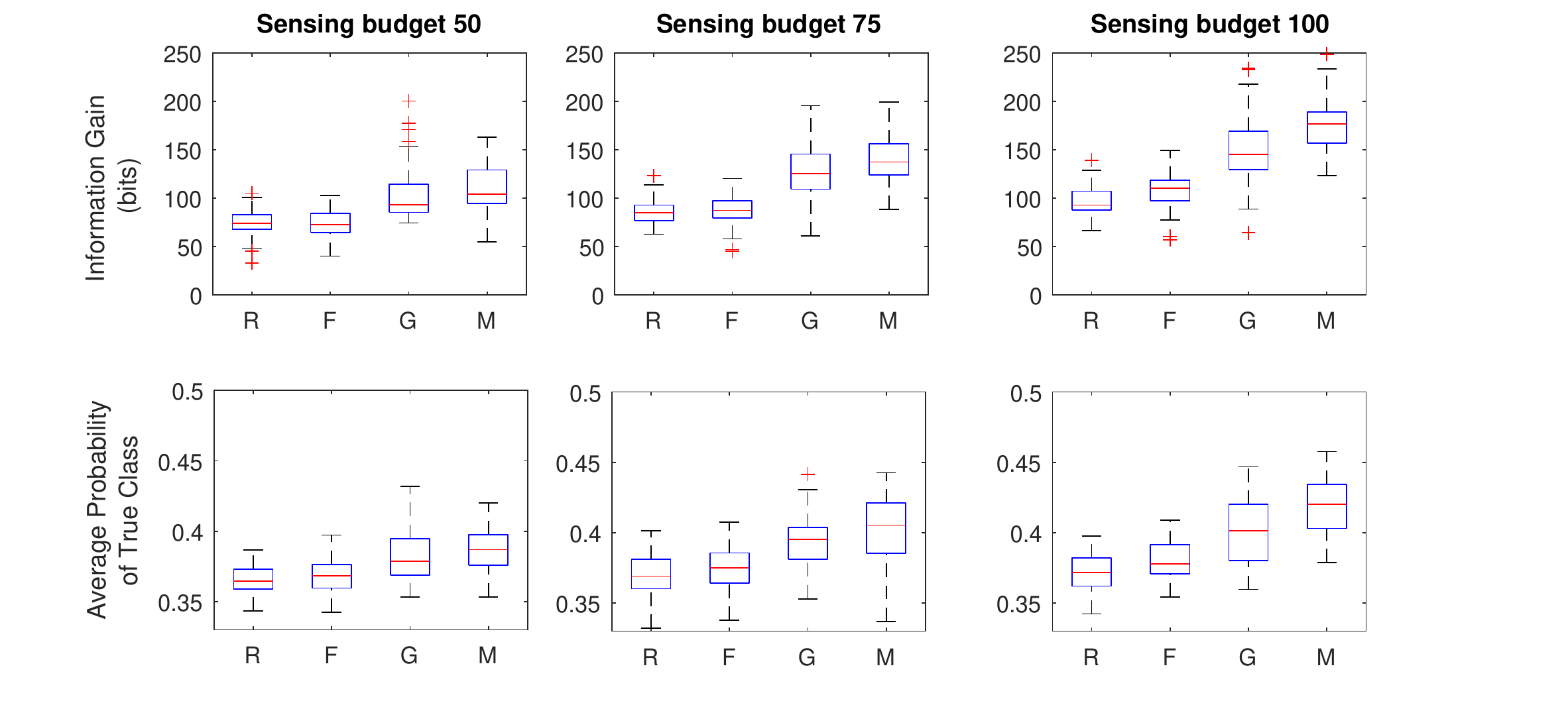}
\caption{Box and whisker plots of information gain and average posterior probability of true class (recognition score) for random policy (R), fixed policy (F), greedy policy (G) and MCTS (M) at different sensing budgets.}
\label{fig:boxplots}
\end{figure*} 

Geologic location types typically occur at a larger spatial scale than rocks. To reflect this, the rock grid was set to be of size $640 \times 640$ where 1.5\% of the cells were randomly selected to contain a rock. Each location grid cell therefore contains six rocks on average. We use three features to characterize a rock. For the purposes of simulation, the actual features used are irrelevant. The UV grid was the same resolution as the location grid.

The domain knowledge BN conditional probability parameters used in the simulation were set such that the camera sensor observes a large area but only observes features weakly associated with the variable of interest while the UV sensor observes a smaller area but takes highly informative measurements. The cost function values of 1 unit and 8 units ensured that good performance is only achieved when both sensors are used in combination.  

All nodes were given ground truth values by randomly sampling from the probability distributions specified in the domain knowledge BN. The robot starts with a uniform distribution of the true value of nodes and this belief is refined as observations are taken. Similarly, the positions of the rocks in the map are also initially unknown to the robot and only discovered as the cells are seen by the camera sensor. The camera sensor can make observations in the rock grid with a rectangular field of view of size $50\times 40$.  The UV sensor observes the UV node in the currently occupied cell. Nodes in the network are assumed to come from one of three categories. 

With these map settings, the space of states the environment could be in is on the order of $10^{400000}$ (joint space of location, rock, feature, and UV grids), and the observation space for a single sensing action with a camera is on the order of $10^{2000}$. Dealing with such large spaces is beyond the capabilities of state of the art POMDP solvers, and further motivates the need for approximate planners.  

We ran trials for random, fixed, greedy and MCTS policies in fifty randomly generated environments and start locations. The policies were tested with sensing budgets of 50, 75 and 100 units. Two performance measures were used: the total information gained and a recognition score, which is defined as the average probability of the correct location class in the robot's belief. For example, if a robot's belief about the class of $L$ in a particular cell is $[0.1,0.2,0.7]$ and the true class is the second one, the ``accuracy" for the cell would be $0.2$. The recognition score is the average accuracy of all of the cells. It is an important metric because it captures situations in which the robot's belief converges to the wrong class. 

The information gain and recognition scores for each policy on the fifty maps are plotted in Fig.~\ref{fig:boxplots}. We carried out a paired t-test to report statistical significance where the pairings were between policies run on the same map instantiations and start positions. Cohen's effect size $d$ is also reported, which is a statistical measure of the performance gap between algorithms. Negative values of $d$ indicate that the performance of the proposed algorithm is greater than the compared algorithm. The magnitude of $d$ gives the size of the effect, with $d > 0.2$, $d > 0.5$ and $d > 0.8$ being thresholds for small, medium and large effects respectively. These are shown in Tables~\ref{infoPval} and~\ref{recogPval}. 

It can be seen visually in the box plot that for all budget sizes, the active algorithms (greedy and MCTS) outperform random and fixed sampling paths in both information gain and recognition score. The paired t-test scores show that MCTS outperforms both random and fixed with p-values $<0.0001$ and large effect sizes. Greedy also statistically significantly outperformed the passive algorithms but this is not shown in the tables here. 

The greedy algorithm did not perform significantly worse than MCTS for a sensing budget of $50$, but starts to drop in performance as the sensing budget is increased. This is because, despite the open and unconstrained nature of the simulation environment, the myopic properties of the greedy algorithm eventually trap it in local minima, a property MCTS is able to better avoid. In a real world unstructured environment with obstacles, where the robot often has to travel through narrow passages to reach high reward regions, we expect MCTS to have even greater performance benefits over a greedy approach. Considering greedy algorithms are popular in many practical implementations of information gathering field robots~\ref{subsec:infoplanning}, this result shows promise that the MCTS approach has potential to greatly improve the performance of field robots in a large range of applications.  

In terms of computation time, each iteration of MCTS took between 0.2 to 0.5 seconds on an average desktop computer. The implementation was however in MATLAB and can be significantly sped up through more efficient memory management and data structures. MCTS also has the advantage of being parallelizable so utilizing multi-threading is also a possibility.

\begin{table}[t]
\centering
\caption{P-values and Cohen's $d$ effect sizes comparing information gain performance for varying budgets. More negative values of $d$ indicate greater performance gap relative to MCTS}
\label{infoPval}
\resizebox{0.45\textwidth}{!}{%
\begin{tabular}{c|cccccc}
\cline{2-7}
 & \multicolumn{6}{c|}{\textbf{Policy}} \\ \hline
\multicolumn{1}{|c|}{\textbf{\begin{tabular}[c]{@{}c@{}}Sensing\\  Budget\end{tabular}}} & \multicolumn{2}{c|}{\textbf{Random}} & \multicolumn{2}{c|}{\textbf{Fixed}} & \multicolumn{2}{c|}{\textbf{Greedy}} \\ \hline
\multicolumn{1}{|c|}{} & \multicolumn{1}{c|}{p} & \multicolumn{1}{c|}{d} & \multicolumn{1}{c|}{p} & \multicolumn{1}{c|}{d} & \multicolumn{1}{c|}{p} & \multicolumn{1}{c|}{d} \\ \hline
\multicolumn{1}{|c|}{50} & 2e-14 & -1.7255 & 5e-12 & -1.7673 & 0.4687 & -0.1507 \\ \cline{1-1}
\multicolumn{1}{|c|}{75} & 7e-16 & -2.3343 & 8e-15 & -2.2135 & 0.0048 & -0.4767 \\ \cline{1-1}
\multicolumn{1}{|c|}{100} & 4e-24 & -3.5713 & 4e-20 & -2.9864 & 0.0005 & -0.7970 \\ \cline{1-1}
\end{tabular}%
}
\end{table}

\begin{table}[t]
\centering
\caption{P-values and Cohen's $d$ effect sizes comparing recognition scores for varying budgets. More negative values of $d$ indicate greater performance gap relative to MCTS}
\label{recogPval}
\resizebox{0.45\textwidth}{!}{%
\begin{tabular}{c|cccccc}
\cline{2-7}
 & \multicolumn{6}{c|}{\textbf{Policy}} \\ \hline
\multicolumn{1}{|c|}{\textbf{\begin{tabular}[c]{@{}c@{}}Sensing\\  Budget\end{tabular}}} & \multicolumn{2}{c|}{\textbf{Random}} & \multicolumn{2}{c|}{\textbf{Fixed}} & \multicolumn{2}{c|}{\textbf{Greedy}} \\ \hline
\multicolumn{1}{|c|}{} & \multicolumn{1}{c|}{p} & \multicolumn{1}{c|}{d} & \multicolumn{1}{c|}{p} & \multicolumn{1}{c|}{d} & \multicolumn{1}{c|}{p} & \multicolumn{1}{c|}{d} \\ \hline
\multicolumn{1}{|c|}{50} & 6e-14 & -1.7705 & 3e-08 & -1.3811 & 0.2540 & -0.2358 \\ \cline{1-1}
\multicolumn{1}{|c|}{75} & 5e-16 & -1.7523 & 3e-09 & -1.5100 & 0.0101 & -0.4757 \\ \cline{1-1}
\multicolumn{1}{|c|}{100} & 3e-19 & -2.6683 & 4e-16 & -2.1761 & 0.0007 & -0.7900 \\ \cline{1-1}
\end{tabular}%
}
\end{table}

\subsection{Planetary Rover Experiments}

\begin{figure*}[!t]
\centering
\includegraphics[width=\textwidth]{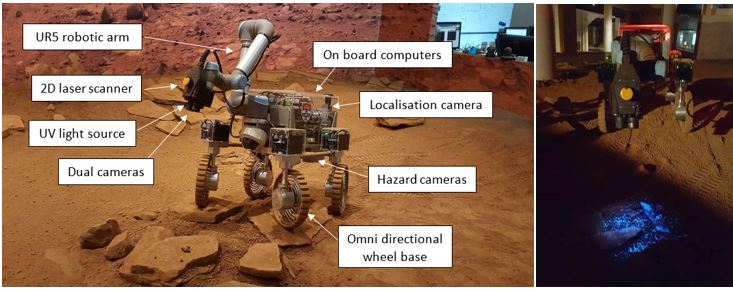}
\caption{Left: System diagram of Continuum. Right: Continuum's UV light source in action}
\label{continuum_diag}
\end{figure*} 

We now demonstrate the practicality of our approach by implementing the BN knowledge network and MCTS planner on a prototype rover and evaluating performance on an analog Martian terrain based in the Museum of Applied Arts and Sciences~(MAAS) in Sydney, Australia. This section summarizes the platform capabilities, the testing environment, our computer vision rock segmentation and feature extraction technique, and presents experimental results.

\subsubsection{Platform Details}
Our rover, Continuum, is pictured in Fig.~\ref{continuum_diag}. It is equipped with an omni-directional drive, which gives it relatively unconstrained motion capabilities. The spiral shape of the rims act as shock absorbers while the rocker arms allow the rover to climb over steep rocks and minimize the changes in orientation. Continuum has a 6-degree-of-freedom robotic arm with cameras, an ultraviolet light source and a 2D laser scanner mounted on the end effector. There are also several hazard cameras around the body to check for collisions. In this experiment we use one of the arm cameras and the UV light source as our two sensors. The light source illuminates the UV reflective powder we discuss in the next section and simulates the role of a spectrometer in a real mission. The arm camera was pointed towards the ground in a pose similar to Fig.~\ref{fig:continuum} to constrain the information that can be gathered in a single observation.

\begin{figure*}[t]
\centering
\includegraphics[width=\textwidth]{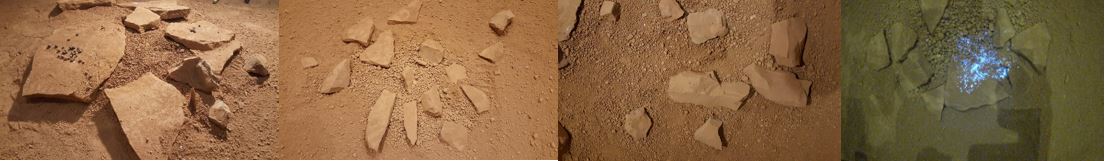}
\caption{From left to right: The three classes of location type and a typical image when the UV light source is activated.}
\label{area_example}
\end{figure*}

\begin{figure*}[t]
\centering
\includegraphics[width=\textwidth]{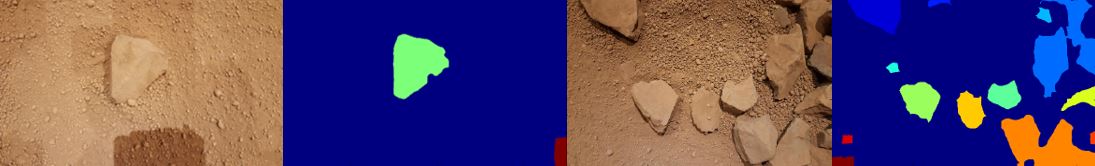}
\caption{Our rock segmentation technique in action. It can be seen that there are false positives in areas with shadows.}
\label{cv_example}
\end{figure*}

\subsubsection{Environment Setup}
Our testing environment, the MAAS Mars Lab, is a $20\times 7$m space that is designed to be a scientifically accurate representation of Martian terrain. The lab was divided into three different types of locations shown in Fig.~\ref{area_example}. Each location type had slightly different distributions of types of rocks and the features they exhibit. UV-reflective powder was added in varying quantities to each category. There was, however, enough ambiguity between categories to encourage the robot to use a combination of both sensors to gather information. 

The rock grid was set to a resolution of 2cm per cell. Rocks are different sizes so they usually span many cells. To account for this we interpret their location to be in the cell nearest to their centroid. The conditional probability parameters of the BN were determined from approximately counting the types of rocks present in the lab and therefore were not 100\% accurate. There were also rocks in the environment which were not explicitly modeled in the BN, which is a realistic source of noise not present in the simulations. In an actual mission, we expect these parameters to be initially determined through domain knowledge and manually tuned by scientists during a communication cycle as data is observed. 

There were also areas in the yard which were dangerous for the robot to traverse such as walls, supporting structures and large rocks. Some rocks were only safely traversable when approached from a specific angle and speed. However, since traversability planning is nontrivial~\cite{peynot2014learned} and outside the scope of this paper, an occupancy map was created to account for the obstacles and given to the robot prior to the mission. However, once the obstacles were sufficiently dilated to account for rover size and localization errors, only a subset of the environment was accessible by the robot. Traversability estimation is therefore essential on a real mission to enable a larger space of sampling paths. Some traversal safety information can be incorporated into our planner by modifying the cost function but extending our approach for more complex dynamic costs is left as future work. 

\subsubsection{Computer vision}
In a realistic unstructured environment the feature extraction process is more complex and requires first segmenting the rocks from the image.  It can be seen in Figs.~\ref{continuum_diag}-\ref{cv_example} that rocks look very similar to ground in terms of color. There are also lighting variations and shadows which complicate the image processing step. There are several methods proposed in the literature which achieved good results. Edge-based techniques such as~\cite{thompson2007performance} ran a Canny edge detector followed by a complex process of pruning and joining edges likely to belong to a rock. Texture based techniques such as~\cite{song2008automated} utilized multi-resolution histograms to achieve coarse segmentation followed by an active contour technique to get good edge detection performance. Another interesting and effective approach was used by~\cite{dunlop2007multi} which calculated superpixels at different scales followed by adding, subtracting, splitting and merging superpixels to satisfy criteria learned from a Support Vector Machine. However, all of these approaches were designed for Martian imagery which did not have the same characteristics as our environment and were not openly available. Furthermore, computation time was not a design factor in these studies so the algorithms often took several minutes to yield a result. 

We approach this problem by first over-segmenting the image into superpixels using the SLIC algorithm~\cite{achanta2012slic} that groups similarly colored pixels together while preserving the strong edges. This is followed by adaptive normalization to reduce lighting variations and shadows. Histograms of intensity, the number of edges, LAB color and intensity variance were calculated for each superpixel and compared to a training image of the ground with no rocks. Applying appropriate thresholds allows us to classify most of the superpixels as rock, ground or shadow. For the more uncertain superpixels, the amount of texture correlation with their local neighborhoods was measured followed by a voting process. This two stage process yields the final image shown in Fig.~\ref{cv_example}. Segmentation is sometimes noisy like most robotic applications, especially in the presence of shadows, but the probabilistic nature of Bayesian networks helps minimize the resulting effects on decision making. For features we use circularity, size and color as they are simple to calculate and geologically meaningful. The UV measurement was obtained by calculating the blue-to-red ratio of the RGB channels. The features and UV measurements were both thresholded into low, medium and high categories. 

\subsubsection{Localization and control}
PID controllers were used in conjunction with a localization system detailed in previous work~\cite{potiris2014terrain} to control the omni-directional drive such that the required position and orientation is achieved within a small error margin. Localization was fused with the computer vision to register observations on a map which allowed the belief space to be updated. The action space was once again discretized into ten actions where the robot could select one of two sensors and decide whether to move forward one step, move diagonally at -45 and 45 degrees or rotate by -90 or 90 degrees. The robot also checked if actions will lead to collisions or cause the robot to drive over valuable rocks through an occupancy map provided to the robot prior to the mission. In a real mission, we expect the localization to come from visual odometry or a full-scale SLAM algorithm.  

\subsubsection{Results} 
We compared our non-myopic planner against a random action policy with random start locations and orientations in the yard. Ten trials were run for each policy. A sensing budget of 30 units was used with a cost function of 1 and 5 units for the camera and UV sensor respectively. We also attempted to implement a greedy strategy but found early in the trials that the robot often got stuck in local minima and was not able to give useful results. This is because, unlike the simulation environment, the presence of obstacles led to narrow paths in the environment that may not be visible with a 1-step horizon. A random policy was able to recover from such situations, gave better results than greedy and hence was a better benchmark. The information gain and recognition scores along with standard deviations are shown in Table~\ref{tab:MCTS_real}. 

\begin{table}[!t]
\centering
\caption{Performance comparison of MCTS planner with random for real robot experiments}
\label{tab:MCTS_real}
\begin{tabular}{|c|c|c|ll}
\hline
\textbf{Policy} & \textbf{Information Gain} & \textbf{Recognition Score}   \\ \hline
Random    & 52.23 (11.76)  & 0.3873 (0.0203)   \\ \hline
MCTS-50   & 59.17 (18.63)  & 0.4068 (0.0255)   \\ \hline
\end{tabular}
\end{table}

If the belief of $L$ were uniform, the accuracy score would be $0.33$. The MCTS active approach therefore gives almost a $25\%$ increase in accuracy score over random policies and $13\%$ increase in terms of information gain. It is important to note the testing environment was relatively small. Longer-horizon plans are likely to generate even more performance benefits as seen in the simulations. 

\begin{figure*}[t]
\centering
\includegraphics[trim = {2.5cm 4cm 2.5cm 2cm},clip,width=\textwidth]{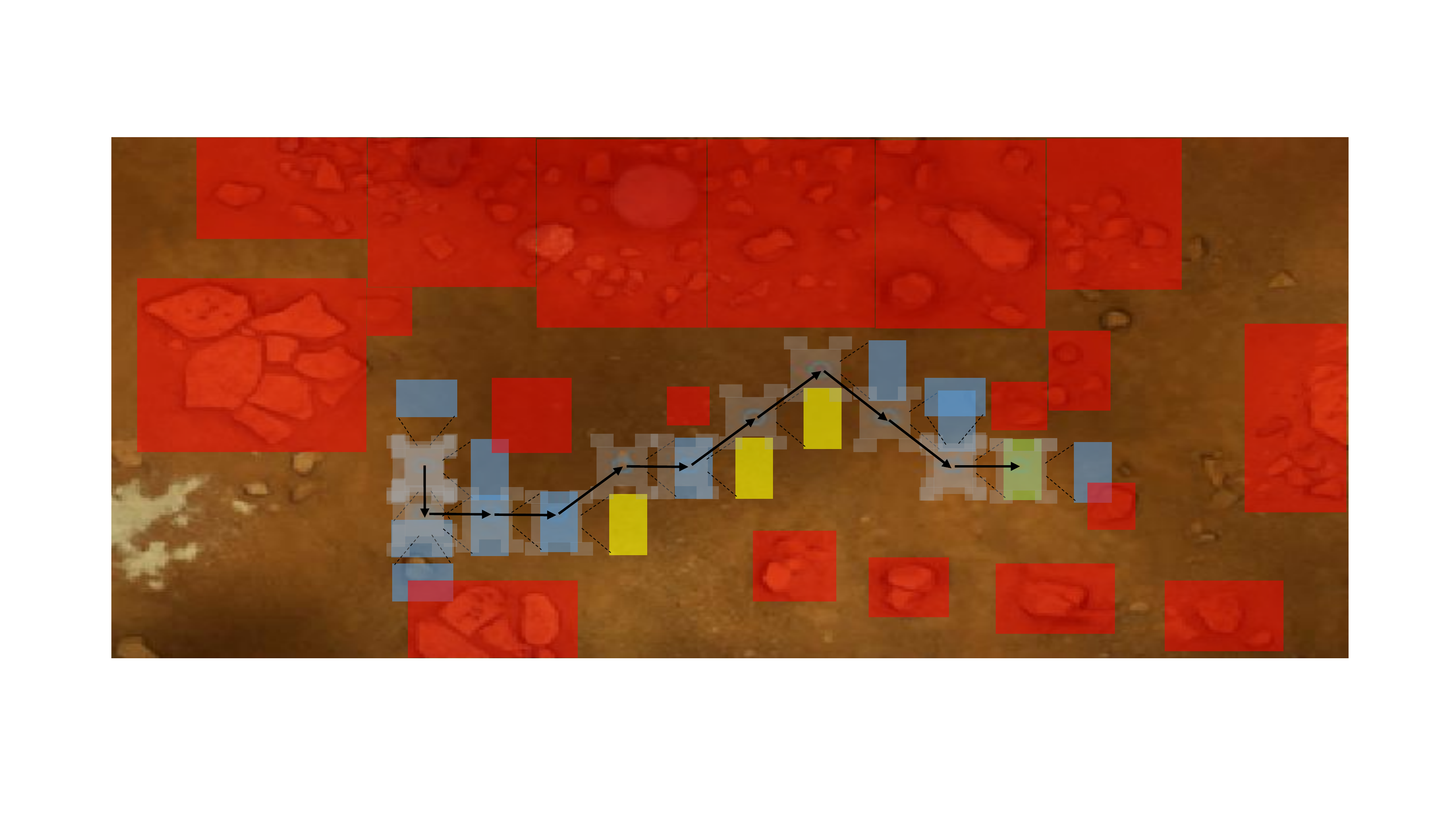}
\caption{An example path followed by the Continuum rover is shown by the black arrows. The rover footprint is in white, whereas the fields of view of the sensors are shown in blue and yellow rectangles. Blue indicates that the camera sensor was used, yellow indicates the UV light source was used, and the green rectangle represents both sensors being used. The red areas are not traversable by the robot.}
\label{fig:examplepath}
\end{figure*}

An example path followed by the robot under the MCTS action selection policy is shown in Fig.~\ref{fig:examplepath}. The path avoids untraversable regions and achieves high spatial coverage of unseen areas in the environment. 

\section{Overcoming incomplete prior knowledge}
\label{sec:mvp}
Often in remote science missions, scientists have incomplete prior knowledge of the key mission variables and parameters. An example is the Mojave Volatiles Prospector~(MVP) project~\cite{heldmann2015mojave}, conducted by NASA Ames Research Center in the Mojave Desert in 2014 and described earlier in Sec.~\ref{sec::introduction}. The purpose of the MVP project was to estimate abundance of subsurface water in a large geographic locale by actively directing a rover equipped with a Neutron Spectrometer System, Near Infra Red Spectrometer and several cameras as seen earlier in Fig.~\ref{fig:continuum}. There was a large team of scientists and engineers working remotely from the testing site who analyzed the sensing data being streamed by the robot and planned future way-points to traverse through (also shown earlier in Fig.~\ref{fig:continuum}). 

The NSS has a small field-of-view, so measurement requires the robot to drive slowly to avoid spatial blurring of readings. The NSS is therefore more expensive in terms of mission time compared to a camera. At the end of the MVP project, the sensor data was analyzed and it was determined that there was a relationship between the albedo of the terrain and the corresponding NSS readings~\cite{foil2016}. If the relationship between the sensing modalities was learned during the mission, scientists could have made more resource-efficient decisions regarding where to drive the robot and deploy sensors to maximize understanding of subsurface water distribution. 

The MVP project was a precursor to the planned Resource Prospector~(RP) project which aims to deploy a robot with a similar sensor suite on the moon and map the abundance of surface volatiles~\cite{andrews2014introducing}. Learning sensor correlations online in combination with our active planning approach will greatly increase the science return of the RP mission, a significant boon given that the project is limited to one lunar day of operations and communication delays are large enough to make direct teleoperation inefficient and unsafe. 

In this section, we study a simulated mission analogous to the MVP project and show how incomplete or imperfect prior knowledge can be overcome by modeling parameter uncertainties in the BN as Dirichlet distributions. We compare our active approach with passive non-adaptive approaches in simulation and with real data from the MVP project and show statistically significant performance benefits.  

\subsection{Scenario Definition}
The operating environment is discretized into a grid where the robot is required to estimate the abundance of water $W$ in each grid cell $n$. While the robot can be equipped with an arbitrary number of sensors, for ease of illustration we consider the case with two sensors: a camera to classify terrain in a cell and a NSS that returns counts that are positively correlated with water abundance. 

Similar to the Mars exploration scenario in Sec.~\ref{sec::marsexploration}, the robot plans action sequences $a_{seq}$ to maximize the expected information gained $EI$ on the water distribution in each cell. The robot must also reach a goal position $x_{goal}$ before it exhausts the sensing budget $S$. The optimization objective is:
\begin{equation}\label{eqn:constraint-problem}
\begin{split}
a_{seq}^{*} &= \operatorname*{arg\,max}_{a_{seq} \in A} EI(a_{seq}) \\
&\textbf{s.t.} \ \textnormal{cost}(a_{seq}) \leq S \\
&\textbf{s.t.} \ x_{end}(x_{start},a_{seq}) = x_{goal}\,.
\end{split}
\end{equation}

In this simulated mission, the camera always takes measurements but the robot must actively decide when to use the NSS. While any motion model can be used, we define the action space $A$ such that the robot can either stay in the current cell and use its NSS or move along one of the four cardinal directions. Any general cost function can be used but for simplicity we use a constant function where movement actions cost 1 unit while using the NSS costs 5 units. 

The expected information gain is given by Eq.~\ref{eqn:info-gain}:
\begin{equation} 
\begin{split}
EI(a_{seq}) &= \sum_{n=1}^{N}{\left[H(W_{n}) - H(W_{n}|a_{seq})\right]} \\
&=  \sum_{n=1}^{N}{\left[H(W_n) - \sum_{Z_{1:L}}{H(W_{n}|Z_{seq})P(Z_{seq}|a_{seq})}\right]}\,,
\end{split}
\label{eqn:info-gain}
\end{equation}
where $H$ is the Shannon entropy, $W_{n}$ is the water abundance in a cell $n$ and $N$ is the total number of cells in the environment. Each action $a_i$ produces a stochastic observation $Z_i$ that reveals information about the water distribution. Term $P(Z_{1:L}|a_{1:L})$ is the sensor noise model and $H(W_{n}|Z_{1:L})$  is a mapping from observable data to subsurface water distribution.
 
\subsection{MVP knowledge modeling}
\begin{figure}[t!]
\centering
\scalebox{0.9}{
\begin{tikzpicture}
\tikzstyle{main}=[circle, minimum size = 10mm, thick, draw =black!80, node distance = 5mm]
\tikzstyle{connect}=[-latex, thick]
  \node[main] (alpha) {$\alpha$};
  \node[main] (theta) [left=of alpha] {$\theta$};
  \node[main] (water) [left=8mm of theta] {$W_n$};
  \node[main, fill = black!10] (NSS) [below=of water] {$Z_{S_n}$};
  \node[main] (terrain) [left=of water] {$T_n$};
  \node[main, fill = black!10] (image_node) [below=of terrain] {$Z_{I_n}$};
  \path (alpha) edge [connect] (theta)
        (theta) edge [connect] (water)
		(water) edge [connect] (NSS)
        (terrain) edge [connect] (image_node)
        (terrain) edge [connect] (water);
  \node[rectangle, inner sep=0mm, fit= (terrain) (NSS),label=below right:N, xshift=-1mm, yshift=1mm] {};
  \node[rectangle, inner sep=4mm, draw=black!100, fit = (terrain) (NSS)] {};
\end{tikzpicture}
}
\caption{Bayesian scientific knowledge model for MVP} 
\label{fig:specific-bn}
\end{figure}
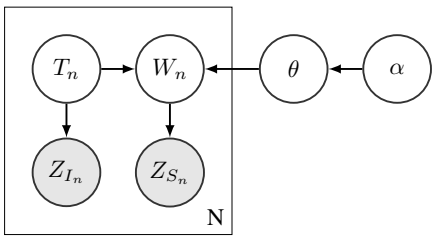

Inspired by topic modeling~\cite{blei2003latent}, we structure the dependencies between the NSS observations, the camera and the subsurface water distribution according to the generative model shown in Fig.~\ref{fig:specific-bn}. The NSS observes the water distribution $W$ in a cell $n$ through observations $Z_{S}$. The camera observation is denoted by $Z_I$ while $T$ is the class of terrain. 

We assume that all nodes are discrete variables but the observation nodes can also represent continuous data. The probabilistic mapping from $T$ and $W$ nodes to their corresponding observation nodes (the $P(Z_I|T)$ and $P(Z_S|W)$ terms) is deduced from the sensor/classifier model. Unsupervised dimensionality reduction techniques can also be applied.

Like in MVP, the probabilistic relationship between terrain and water classes is unknown at the beginning of the mission, an example of incomplete scientific knowledge. We parametrize this relationship by vector $\theta$ which is modeled as a Dirichlet distribution with hyperparameters $\alpha$. The hyperparameters are updated during the mission as data is collected. This allows the robot to automatically learn and refine any correlations that may be present and adapt its decision-making accordingly. More formally,
\begin{equation}\label{eq:pwtdef}
P(W|T=t) \sim Categorical(\theta_t)\,,
\end{equation}
\begin{equation}
\theta_t \sim Dirichlet(\mathbf{\alpha_t})\,,
\end{equation}
\begin{equation}
\theta = [\theta_1,\theta_2\dots\theta_T]\,.
\end{equation}

\subsection{Updating Node Beliefs}
In this section we derive the Bayesian update equations for the beliefs of nodes $W_{n}$, $T_{n}$ and $\theta$ as observations are made in a cell. We begin with the belief update equations for water abundance in the general case where both NSS and image observations are taken. For compactness, the subscript $n$ in terms $W_{n}$ and $T_{n}$ is dropped. Applying Bayes' theorem we get the following equation where $\eta$ is the normalization constant:
\begin{equation}
P(W|Z_I, Z_S) =\eta P(Z_S|W)P(W|Z_I)\,. 
\end{equation}
Expanding the $P(W|Z_I)$ term:
\begin{equation}
\begin{split}
P(W|Z_I) &=\sum_{T}{P(T|Z_I)P(W|Z_I, T)} \\
&=\eta\sum_{T}{P(T)P(Z_I|T)P(W|Z_I, T)} \\
&=\eta\sum_{T}{P(T)P(Z_I|T) \int_{\theta} P(W|T,\theta)P(\theta) d\theta}\,. 
\end{split}
\end{equation}
Applying Eq.~\ref{eq:pwtdef}:
\begin{equation}
\begin{split}
P(W|Z_I) &=\eta\sum_{T}{P(T)P(Z_I|T) \int_{\theta} \theta P(\theta) d\theta} \\
&=\eta\sum_{T}{P(T)P(Z_I|T) \mathbb{E}(\theta)}\,.
\end{split}
\end{equation}
Similarly, we can iteratively update belief on terrain type by evaluating
\begin{equation}
P(T|Z_I, Z_S) = \eta P(T)P(Z_I|T) \sum_{W}{P(Z_S|W) \mathbb{E}(\theta)}\,.
\end{equation}
Since $\theta$ is modeled by a Dirichlet distribution, $\mathbb{E}(\theta)$ can be efficiently calculated by normalizing the corresponding hyperparameters
\begin{equation}
\mathbb{E}(\theta_{wt}) = P(W = w | T = t) = \frac{\alpha_{wt}}{\sum_{k}^{|W|}{\alpha_{kt}}}\,.
\end{equation}
Lastly we can update $\theta$ with
\begin{equation}
\begin{split}
P(\theta|\alpha, Z) &= \sum_{T,W}{P(\theta|\alpha_{init}, Z, T, W)P(T,W|Z)} \\
&= \sum_{T,W}{P(\theta|\alpha_{init}, T, W)P(T,W|Z)}\,,
\end{split}
\end{equation}
where $Z$ is a compact form of the full observation vector $[Z_I, Z_S]$.

Dirichlet distributions and categorical distributions are conjugate priors. This means that $P(\theta|\alpha, Z)$ is also a Dirichlet distribution and we can calculate the posterior by simply updating the Dirichlet hyperparameters by evaluating $\alpha_{w,t} = \alpha_{w,t} + P(W = w, T = t|Z)$ for all values of $W$ and $T$, where $w \in \{1,\ldots,|W|\}$ and $t \in \{1,\ldots,|T|\}$. When $|W|$ and $|T|$ become large, Gibbs sampling approaches can be used to approximate this update~\cite{girdhar2016modeling}. When a terrain cell is observed we also update the terrain beliefs in neighboring cells using a Gaussian kernel in a similar fashion to Fig.~\ref{knowledgebn}.

\begin{table*}[t]
\centering
\caption{P-values and Cohen's $d$ effect sizes comparing information gain performance for varying budgets. More negative values of $d$ indicate greater performance gap relative to MCTS \label{tbl:mvpinfo-gain}}
\begin{tabular}{c|cccc|cccc|cccc|cc}
\hline
\multirow{2}{*}{\textbf{Budget}} &  \multicolumn{4}{c}{\textbf{Greedy}} & \multicolumn{4}{c}{\textbf{Random}} & \multicolumn{4}{c}{\textbf{Lawnmower}} & \multicolumn{2}{c}{\textbf{MCTS-50}} \\
& $\mu$ & $\sigma$ & $p$ & $d$ & $\mu$ & $\sigma$ & $p$ & $d$ & $\mu$ & $\sigma$ & $p$ & $d$ & $\mu$ & $\sigma$ \\
\hline
60 & 20.7 & 8.93 & \textbf{0.05} & \textbf{-0.34} & 15.6 & 7.16 & \textbf{1e-5} & \textbf{-0.95} & 22.4 & 8.99 & 0.40 & -0.17 & 24.0 & 10.23 \\
80 & 28.4 & 11.76 & \textbf{0.003} & \textbf{-0.60} & 20.6 & 10.92 & \textbf{9e-8} & \textbf{-1.20} & 31.7 & 12.62 & \textbf{0.03} & \textbf{-0.35} & 36.7 & 15.54 \\
100 & 32.3 & 12.81 & \textbf{0.004} & \textbf{-0.57} & 27.1 & 13.42 & \textbf{1e-4} & \textbf{-0.88} & 39.4 & 14.50 & 0.52 & -0.12 & 41.4 & 18.81 \\
120 & 39.3 & 13.99 & \textbf{0.003} & \textbf{-0.62} & 29.0 & 12.93 & \textbf{2e-8} & \textbf{-1.31} & 46.6 & 19.24 & 0.39 & -0.14 & 49.1 & 17.37 \\
140 & 43.4 & 13.12 & \textbf{3e-5} & \textbf{-0.84} & 33.3 & 16.58 & \textbf{5e-9} & \textbf{-1.34} & 54.8 & 21.4 & 0.62 & -0.10 & 56.8 & 18.50 \\
\hline
\end{tabular}
\end{table*}

\begin{table*}[t]
\centering
\caption{The average posterior probability of the true water distribution given the maps learned by the different algorithms. Larger values are desirable.
\label{tbl:mvpacc-score}}
\begin{tabular}{c|cccc|cccc|cccc|cc}
\hline
\multirow{2}{*}{\textbf{Budget}} &  \multicolumn{4}{c}{\textbf{Greedy}} & \multicolumn{4}{c}{\textbf{Random}} & \multicolumn{4}{c}{\textbf{Lawnmower}} & \multicolumn{2}{c}{\textbf{MCTS-50}} \\
& $\mu$ & $\sigma$ & $p$ & $d$ & $\mu$ & $\sigma$ & $p$ & $d$ & $\mu$ & $\sigma$ & $p$ & $d$ & $\mu$ & $\sigma$ \\
\hline
60 & 0.37 & 0.02 & 0.56 & -0.08 & 0.36 & 0.02 & \textbf{0.004} & \textbf{-0.43} & 0.38 & 0.02 & 0.14 & 0.25 & 0.38 & 0.03 \\
80 & 0.39 & 0.03 & \textbf{0.008} & \textbf{-0.44} & 0.37 & 0.03 & \textbf{3e-8} & \textbf{-0.94} & 0.40 & 0.03 & 0.17 & -0.19 & 0.41 & 0.04 \\
100 & 0.41 & 0.03 & 0.21 & -0.19 & 0.38 & 0.03 & \textbf{8e-5} & \textbf{-0.81} & 0.42 & 0.03 & 0.36 & 0.15 & 0.41 & 0.05 \\
120 & 0.42 & 0.04 & \textbf{0.04} & \textbf{-0.35} & 0.38 & 0.04 & \textbf{5e-10} & \textbf{-1.37} & 0.43 & 0.03 & 0.84 & -0.03 & 0.43 & 0.03 \\
140 & 0.43& 0.04 & \textbf{0.0008} & \textbf{-0.54} & 0.39 & 0.04 & \textbf{3e-12} & \textbf{-1.62} & 0.44 & 0.04 & 0.15 & -0.26 & 0.45 & 0.03 \\
\hline
\end{tabular}
\end{table*}

\subsection{Simulation analysis}

In this section we evaluate the performance of the MCTS approach under these new conditions against alternative approaches. As mentioned in Sec.~\ref{sec::relatedwork}, there are several algorithms in literature for informative path planning \cite{hollinger2014sampling,binney2012branch} but these approaches are not suitable when multiple sensors are involved and planning has to occur in an initially unknown environment. We therefore compare the performance of our approach with the following three baseline algorithms:

\noindent \textbf{Random:} At each time step the robot determines the set of actions it can execute in the next step without breaking the goal position and sensing budget constraints. This is determined by calculating the lowest cost paths from candidate nodes to the goal using A* and checking whether they are feasible with the remaining sensing budget. With the action space we used, this was equivalent to calculating the Manhattan distance. A random action is chosen out of this set. The random policy serves as a baseline for algorithm performance.

\noindent \textbf{Greedy:} At each time step, out of the reachable action set, the robot selects the action with the highest expected information gain for the water abundance to sensing cost ratio. This is given by:
\begin{equation}
a_{next}^{*} = \operatorname*{arg\,max}_{a \in A} \frac{\sum_{z}{I(z)P(z|a)}}{cost(a)}.
\end{equation}

\noindent \textbf{Lawnmower:} We use a ``lawnmower" pattern to achieve coverage of the environment. Here we arbitrarily allocate 50\% of the sensing budget to the path and 50\% to using the NSS. A lawnmower-like path is designed manually which adheres to the sensing budget as well as the initial and final positions. The NSS is used at uniform intervals along the path. Any planner which optimizes for spatial coverage rather than the actual expected value of the observations can be expected to have similar performance. 

Our approach, \textbf{MCTS-50} (50 iterations were used for MCTS) was evaluated against the baseline algorithms on 50 randomly generated $20\times 20$ Voronoi maps with fixed start and goal positions. Terrain, water, and the observation nodes were categorical variables with three classes. The true correlation between terrain type and water class (initially unknown to the robot) was set to be $0.85$, i.e., given the terrain class, the water class could be predicted with $85\%$ accuracy. Sensor noise for the terrain was set to be 10\% while the NSS had 5\%. All unobserved nodes were given an uniform prior and the $\alpha$ hyperparameters were initialized to a value of $1$. Two performance metrics were used: information gain and the recognition score, which we defined earlier as the average posterior probability of the correct class of water in the cells. Pairwise t-test scores and Cohen's effect size $d$ are also reported. 

The results are shown in Tables~\ref{tbl:mvpinfo-gain}~and~\ref{tbl:mvpacc-score}. In terms of average information gain, our approach statistically outperform random and greedy policies with notable effect sizes (bolded). For the recognition score, the performance improvement is less pronounced. This is because the robot observes a small proportion of the map and the unseen areas dominate the score.

\begin{table}[t]
\centering
\caption{Information gain and accuracy scores along with t-test scores and effect sizes for Experiment 1. MCTS outperforms spatial coverage in both metrics with medium effect sizes.}
\label{tbl:prior1}
\begin{tabular}{c|c|c|c|c|}
\cline{2-5}
 & Lawnmower & MCTS & $p$ & $d$ \\ \hline
\multicolumn{1}{|c|}{I} & 43.70(15.58) & 53.82(17.91) & 4e-4 & -0.5105 \\ \hline
\multicolumn{1}{|c|}{A} & 0.4472(0.0329) & 0.4741(0.0380) & 2e-5 & -0.5350 \\ \hline
\end{tabular}
\end{table}

\begin{table}[t]
\centering
\caption{Information gain and accuracy scores along with t-test scores and effect sizes for Experiment 2. MCTS outperforms spatial coverage in both metrics with medium to large effect sizes.}
\label{tbl:prior2}
\begin{tabular}{c|c|c|c|c|}
\cline{2-5}
 & Lawnmower & MCTS & $p$ & $d$ \\ \hline
\multicolumn{1}{|c|}{I} & 44.72(14.68) & 65.08(21.77) & 7e-8 & -0.7753 \\ \hline
\multicolumn{1}{|c|}{A} & 0.4263(0.0281) & 0.4704(0.0400) & 2e-10 & -0.9034 \\ \hline
\end{tabular}
\end{table}

In the simulated conditions, the performance of lawnmower is comparable to MCTS. In completely unknown and open environments, paths that provide good spatial coverage of the environment are indeed a logical and effective way to gain information. In more realistic environments with obstacles, planning lawnmower paths becomes more complicated. When environmental obstacles are known a priori, Boustrophedon coverage can be applied~\cite{choset2001coverage}. In unknown or partially known environments, however, additional re-planning subroutines would need to be implemented as obstacles are discovered. 

Further, adapting the lawnmower approach to an arbitrary number of sensors would require a way to split the sensing budget across the different sensing modalities, which our approach optimizes automatically in a principled manner. While the 50-50 budget split between paths and NSS produced good results in the simulation setting, there is no guarantee that performance will continue to be competitive in field settings with longer missions and large environments.

In robotic missions, there is usually some prior information available about the scene such as orbital maps or scientific hypotheses on what the robot is likely to see. A key advantage of our approach is that we can easily encode this knowledge in the form of Bayesian priors. Orbital maps can be encoded by biasing the prior distribution of terrain types while scientific knowledge of known sensor correlations can be incorporated by incrementing the $\alpha$ hyperparameters. Unlike the standard lawnmower, our approach will automatically take advantage of this information without algorithmic modifications. To validate this, we ran two sets of experiments where robot had access to different types of prior knowledge.

\begin{figure*}[ht]
    \centering
    \hfill
    \subfloat[b][Pavement Terrain Type]{
    	\includegraphics[width=0.32\textwidth]{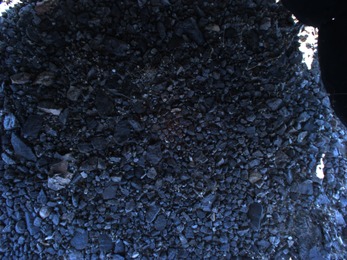}}
    ~ 
      \hfill
    \subfloat[b][Transition Terrain Type]{
        \includegraphics[width=0.32\textwidth]{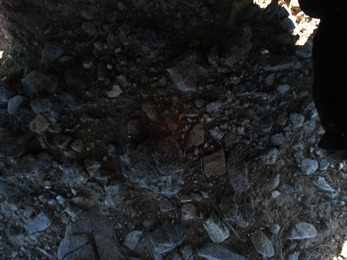}}
    ~ 
    \hfill
    \subfloat[b][Wash Terrain Type]{
        \includegraphics[width=0.32\textwidth]{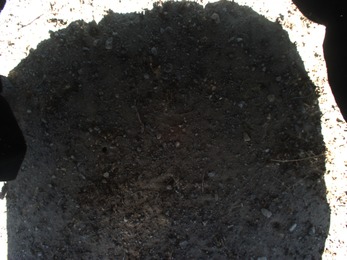}}
    \hfill
    \caption{Different types of terrain in the MVP test area.  Pavements were found to be associated with high NSS counts, while washes had low NSS counts.  The transition terrain was in between washes and pavements and had moderate NSS counts.\label{fig:terrain-types}}
\end{figure*}

\noindent\textbf{Experiment 1:} The robot's belief of the correct terrain type was initialized to $0.5$ instead of a uniform distribution of $0.33$. This is analogous to the information that orbital maps provide. 

\noindent\textbf{Experiment 2:} The Dirichlet hyperparameters for how one particular terrain class maps to the water distribution were initialized to a non-uniform value. This simulates the scenario where scientists have prior knowledge (for example based on geological mechanisms or previous expeditions) on how one class of terrain correlates with water distribution. 

50 trials were run on the same set of maps as the experiments in Tables~\ref{tbl:mvpinfo-gain}~and~\ref{tbl:mvpacc-score} with a sensing budget of 140 units. The results are shown in Tables~\ref{tbl:prior1}~and~\ref{tbl:prior2} where I and A represent information gain and accuracy scores. The standard deviation is given in brackets. MCTS statistically outperformed lawnmower in both sets of experiments with medium and large effect sizes, which suggests that there is a strong advantage in employing our active approach over passive non-adaptive approaches. 

\subsection{Results with real data}
Since much of the data from the Mojave Desert test site was collected in line traverses, we selected 100 pairs of ground camera images and NSS counts from this dataset and redistributed them into a $10\times 10$ grid to simulate a field environment. Typical ground camera images are shown in Fig.~\ref{fig:terrain-types}. The images from the MVP dataset are noisy, with both strong shadows and regions with saturation. 

To transform this data into a representation that is suitable for the BN model, we use a simple example-based classifier for illustration. We selected image subsets based on domain knowledge of the terrain classes present and used these to define four cluster centers. Candidate images are then classified based on the closest cluster center in intensity space. The labels are transformed into soft evidence using a confusion matrix derived from training data. Similarly, k-means clustering with three clusters is used to probabilistically classify NSS counts into water abundance. The probabilistic classifications are input to the BN as soft evidence. Continuous data can also be directly fed into the proposed generative model as long as the probabilistic mapping from $T$ and $W$ nodes to observations can be determined. 

We compared MCTS-50 and lawnmower on 20 randomly generated $10\times 10$ maps with a sensing budget of 40 units. We ran two sets of trials with NSS costs of 2 and 5 units. Since the sensing budget of 40 is relatively small, varying the cost of the NSS artificially changes the planning horizon and intends to show the resulting changes in performance.  

\begin{figure}[t]
\centering
\includegraphics[trim={0.5cm 3.5cm 2.5cm 0.9cm},clip,width=0.45\textwidth]{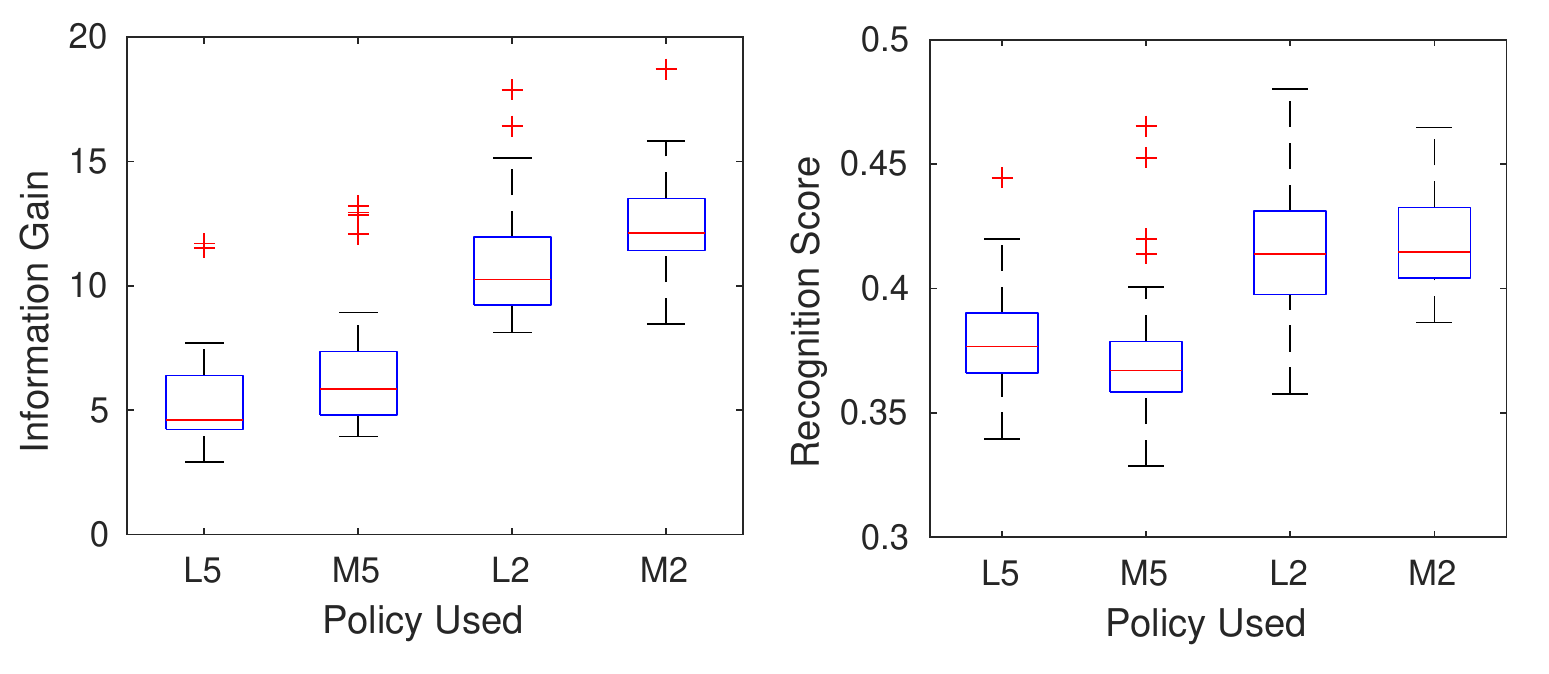}
\caption{Comparison of information gain and recognition scores for lawnmower and MCTS with different NSS costs.\label{fig:real-results}}
\end{figure}

Results are shown in Fig.~\ref{fig:real-results}. In terms of information gain, MCTS is on average better than lawnmower for this sample and the difference is statistically significant when the NSS cost is 2 ($p<0.01$). There is a larger performance gap compared to the simulations because the 50-50 split in the lawnmower budget allocation is no longer as effective for this map size, sensing budget and sensor model. We assumed an initially unknown environment and further improvements can be expected with the integration of prior knowledge. In terms of recognition score, MCTS is slightly lower than lawnmower in NSS-5 and similar in NSS-2, but statistically indifferent.

\section{Conclusions and Future Work}
This paper discussed the multi-modal scientific information gathering problem in which the robot is required to learn about a variable of scientific interest that cannot directly be measured but must be estimated by combining information from multiple sensing modalities with domain knowledge. 
We presented an active perception solution that allows the robot to model and plan with scientists' domain knowledge onboard, which is a powerful new capability for information gathering robots. 

BNs enabled us to model causal and quantitative relationships while robustly handling any uncertainties that may be present, while our adaptation and application of the MCTS technique provided a principled way to plan long horizon informative paths in partially observable environments with multiple sensing modalities in an anytime manner. With appropriate design, the approach also allows recursive updating of key variables and avoids the need to store an history of observations. These properties make our approach highly applicable for real time execution on field robots with limited onboard computational capabilities. Experiments were conducted in simulation, with data from a past expedition, and on a prototype space rover. It was shown that our active approach statistically significantly outperforms passive and myopic approaches popular in similar applications. 

There are several promising areas for future work. Richer knowledge representation frameworks such as statistical relational models could be explored for more complex applications where BNs cannot adequately capture the key relationships and dependencies present. Another interesting line of work is to adapt the structure of the BN online to better fit and predict observations. This would allow robots to automatically discover previously unknown relationships, which could revolutionize science methodology for remote environments.

The computation time for MCTS was dominated by the forward simulation of belief updates to estimate rewards. Research into approximating the belief space or variational approaches to calculate information gain will lead to significant increases in performance and will allow field robots to handle larger, more complex BN structures, and plan for longer horizons in real time. 

\section*{Acknowledgements}
The authors would like to thank Richard Elphic at NASA Ames Research Center for his help in understanding the Neutron Spectrometer System data. 


\bibliographystyle{IEEEtran}      
\bibliography{biblio_ICRA.bib}   

\begin{thebibliography}{10}
\providecommand{\url}[1]{#1}
\csname url@samestyle\endcsname
\providecommand{\newblock}{\relax}
\providecommand{\bibinfo}[2]{#2}
\providecommand{\BIBentrySTDinterwordspacing}{\spaceskip=0pt\relax}
\providecommand{\BIBentryALTinterwordstretchfactor}{4}
\providecommand{\BIBentryALTinterwordspacing}{\spaceskip=\fontdimen2\font plus
\BIBentryALTinterwordstretchfactor\fontdimen3\font minus
  \fontdimen4\font\relax}
\providecommand{\BIBforeignlanguage}[2]{{%
\expandafter\ifx\csname l@#1\endcsname\relax
\typeout{** WARNING: IEEEtran.bst: No hyphenation pattern has been}%
\typeout{** loaded for the language `#1'. Using the pattern for}%
\typeout{** the default language instead.}%
\else
\language=\csname l@#1\endcsname
\fi
#2}}
\providecommand{\BIBdecl}{\relax}
\BIBdecl

\bibitem{grotzinger2012mars}
J.~P. Grotzinger, J.~Crisp, A.~R. Vasavada, R.~C. Anderson, C.~J. Baker,
  R.~Barry, D.~F. Blake, P.~Conrad, K.~S. Edgett, B.~Ferdowski \emph{et~al.},
  ``Mars science laboratory mission and science investigation,'' \emph{Space
  Sci. Rev.}, vol. 170, no. 1-4, pp. 5--56, 2012.

\bibitem{heldmann2015mojave}
J.~Heldmann, A.~Colaprete, A.~Cook, T.~Roush, M.~Deans, R.~Elphic, D.~Lim,
  J.~Skok, N.~Button, S.~Karunatillake \emph{et~al.}, ``Mojave volatiles
  prospector {(MVP)}: Science and operations results from a lunar polar rover
  analog field campaign,'' in \emph{Proc. of Lunar Planet. Sci.}, vol.~46,
  2015, p. 2165.

\bibitem{dunbabin2012robots}
M.~Dunbabin and L.~Marques, ``Robots for environmental monitoring: Significant
  advancements and applications,'' \emph{IEEE Robot. Autom. Mag.}, vol.~19,
  no.~1, pp. 24--39, 2012.

\bibitem{leshin2013volatile}
L.~Leshin, P.~Mahaffy, C.~Webster, M.~Cabane, P.~Coll, P.~Conrad, P.~Archer,
  S.~Atreya, A.~Brunner, A.~Buch \emph{et~al.}, ``Volatile, isotope, and
  organic analysis of {Martian} fines with the {Mars Curiosity} rover,''
  \emph{Science}, vol. 341, no. 6153, p. 1238937, 2013.

\bibitem{andrews2014introducing}
D.~R. Andrews, A.~Colaprete, J.~Quinn, D.~Chavers, and M.~Picard, ``Introducing
  the resource prospector ({RP}) mission,'' in \emph{Proc. of AIAA SPACE},
  2014, p. 4378.

\bibitem{bajcsy1988active}
R.~Bajcsy, ``Active perception,'' \emph{Proc. of IEEE}, vol.~76, no.~8, pp.
  966--1005, 1988.

\bibitem{castano2007oasis}
R.~Castano, T.~Estlin, R.~C. Anderson, D.~M. Gaines, A.~Castano, B.~Bornstein,
  C.~Chouinard, and M.~Judd, ``{OASIS}: {O}nboard autonomous science
  investigation system for opportunistic rover science,'' \emph{J. Field
  Robot.}, vol.~24, no.~5, pp. 379--397, 2007.

\bibitem{estlin2012aegis}
T.~A. Estlin, B.~J. Bornstein, D.~M. Gaines, R.~C. Anderson, D.~R. Thompson,
  M.~Burl, R.~Casta{\~n}o, and M.~Judd, ``{AEGIS} automated science targeting
  for the {MER} opportunity rover,'' \emph{ACM Trans. Intel. Syst. Tec.},
  vol.~3, no.~3, p.~50, 2012.

\bibitem{ma2017data}
K.-C. Ma, L.~Liu, H.~K. Heidarsson, and G.~S. Sukhatme, ``Data-driven learning
  and planning for environmental sampling,'' \emph{J. Field Robot.}, 2017.

\bibitem{das2013hierarchical}
J.~Das, J.~Harvey, F.~Py, H.~Vathsangam, R.~Graham, K.~Rajan, and G.~S.
  Sukhatme, ``Hierarchical probabilistic regression for {AUV}-based adaptive
  sampling of marine phenomena,'' in \emph{Proc. of IEEE ICRA}, 2013, pp.
  5571--5578.

\bibitem{hollinger2014sampling}
G.~A. Hollinger and G.~S. Sukhatme, ``Sampling-based robotic information
  gathering algorithms,'' \emph{Int. J. Robot. Res.}, vol.~33, no.~9, pp.
  1271--1287, 2014.

\bibitem{marchant2014bayesian}
R.~Marchant and F.~Ramos, ``Bayesian optimisation for informative continuous
  path planning,'' in \emph{Proc. of IEEE ICRA}, 2014, pp. 6136--6143.

\bibitem{bender2013autonomous}
A.~Bender, S.~B. Williams, and O.~Pizarro, ``Autonomous exploration of
  large-scale benthic environments,'' in \emph{Proc. of IEEE ICRA}, 2013, pp.
  390--396.

\bibitem{hunter2010survey}
A.~Hunter and W.~Liu, ``A survey of formalisms for representing and reasoning
  with scientific knowledge,'' \emph{The Knowledge Engineering Review},
  vol.~25, no.~02, pp. 199--222, 2010.

\bibitem{pearl2014probabilistic}
J.~Pearl, \emph{Probabilistic reasoning in intelligent systems: {N}etworks of
  plausible inference}.\hskip 1em plus 0.5em minus 0.4em\relax Morgan Kaufmann,
  2014.

\bibitem{roy2005finding}
N.~Roy, G.~Gordon, and S.~Thrun, ``Finding approximate pomdp solutions through
  belief compression,'' \emph{JAIR}, vol.~23, pp. 1--40, 2005.

\bibitem{silver2016mastering}
D.~Silver, A.~Huang, C.~J. Maddison, A.~Guez, L.~Sifre, G.~Van Den~Driessche,
  J.~Schrittwieser, I.~Antonoglou, V.~Panneershelvam, M.~Lanctot \emph{et~al.},
  ``Mastering the game of {Go} with deep neural networks and tree search,''
  \emph{Nature}, vol. 529, no. 7587, pp. 484--489, 2016.

\bibitem{best2016decentralised}
G.~Best, O.~M. Cliff, T.~Patten, R.~R. Mettu, and R.~Fitch, ``Decentralised
  {Monte Carlo} tree search for active perception,'' in \emph{Proc. of WAFR},
  2016.

\bibitem{nguyen2016real}
J.~L. Nguyen, N.~R. Lawrance, R.~Fitch, and S.~Sukkarieh, ``Real-time path
  planning for long-term information gathering with an aerial glider,''
  \emph{Auton. Robot.}, vol.~40, no.~6, pp. 1017--1039, 2016.

\bibitem{foil2016}
G.~Foil, Fong, D.~M. Terry, R.~C. Elphic, and D.~Wettergreen, ``Physical
  process models for improved rover mapping,'' in \emph{Proc. of iSAIRAS},
  2016.

\bibitem{arorairos}
A.~Arora, R.~Fitch, and S.~Sukkarieh, ``An approach to autonomous science by
  modeling geological knowledge in a {Bayesian} framework,'' in \emph{Proc. of
  IEEE/RSJ IROS}, 2017.

\bibitem{aroraFSR}
A.~Arora, M.~Furlong, R.~Fitch, T.~Fong, S.~Sukkarieh, and R.~Elphic, ``Online
  multi-modal learning and adaptive informative trajectory planning for
  autonomous exploration,'' in \emph{Proc. of FSR}, 2017.

\bibitem{ellery2017robotic}
A.~Ellery, ``Robotic astrobiology--prospects for enhancing scientific
  productivity of {Mars} rover missions,'' \emph{Int. J. Astrobiol.}, pp.
  1--15, 2017.

\bibitem{cabrol2007life}
N.~A. Cabrol, D.~Wettergreen, K.~Warren-Rhodes, E.~A. Grin, J.~Moersch, G.~C.
  Diaz, C.~S. Cockell, P.~Coppin, C.~Demergasso, J.~M. Dohm \emph{et~al.},
  ``Life in the {Atacama}: Searching for life with rovers (science overview),''
  \emph{J. Geophys. Res-Biogeo}, vol. 112, no.~G4, 2007.

\bibitem{van2005development}
M.~Van~Winnendael, P.~Baglioni, and J.~Vago, ``Development of the {ESA}
  {ExoMars} rover,'' in \emph{Proc. of iSAIRAS}, 2005, pp. 5--8.

\bibitem{smith2007probabilistic}
T.~Smith, ``Probabilistic planning for robotic exploration,'' Ph.D.
  dissertation, Robotics Institute, Carnegie Mellon University, 2007.

\bibitem{thompson2008intelligent}
D.~R. Thompson and D.~Wettergreen, ``Intelligent maps for autonomous
  kilometer-scale science survey,'' \emph{Proc. of iSAIRAS}, 2008.

\bibitem{thompson2011autonomous}
D.~R. Thompson, D.~S. Wettergreen, and F.~J.~C. Peralta, ``Autonomous science
  during large-scale robotic survey,'' \emph{J. Field Robot.}, vol.~28, no.~4,
  pp. 542--564, 2011.

\bibitem{azmanincorporating}
K.~Azman, ``Incorporating prior knowledge into {G}aussian process models,'' in
  \emph{Proc. of Int. PhD Work. Systems and Control}, pp. 253--256.

\bibitem{lawrence2004gaussian}
N.~D. Lawrence, ``Gaussian process latent variable models for visualisation of
  high dimensional data,'' in \emph{Adv. Neur. In.}, 2004, pp. 329--336.

\bibitem{rasmussen2006gaussian}
C.~E. Rasmussen, \emph{Gaussian processes for machine learning}.\hskip 1em plus
  0.5em minus 0.4em\relax MIT Press, 2006.

\bibitem{knight2001casper}
S.~Knight, G.~Rabideau, S.~Chien, B.~Engelhardt, and R.~Sherwood, ``{CASPER}:
  Space exploration through continuous planning,'' \emph{IEEE Intell. Syst.},
  vol.~16, no.~5, pp. 70--75, 2001.

\bibitem{francis2017aegis}
R.~Francis, T.~Estlin, G.~Doran, S.~Johnstone, D.~Gaines, V.~Verma, M.~Burl,
  J.~Frydenvang, S.~Monta{\~n}o, R.~Wiens \emph{et~al.}, ``{AEGIS} autonomous
  targeting for {ChemCam} on {Mars Science Laboratory}: Deployment and results
  of initial science team use,'' \emph{Science Robot.}, vol.~2, no.~7, p.
  eaan4582, 2017.

\bibitem{chien2004eo}
S.~Chien, R.~Sherwood, D.~Tran, B.~Cichy, G.~Rabideau, R.~Castano, A.~Davies,
  R.~Lee, D.~Mandl, S.~Frye \emph{et~al.}, ``The {EO-1} autonomous science
  agent,'' in \emph{Proc. of Int. Joint Conf. Auton. Agents \& Multiagent
  Syst.}\hskip 1em plus 0.5em minus 0.4em\relax IEEE Computer Society, 2004,
  pp. 420--427.

\bibitem{woods2009autonomous}
M.~Woods, A.~Shaw, D.~Barnes, D.~Price, D.~Long, and D.~Pullan, ``Autonomous
  science for an {ExoMars Rover}--like mission,'' \emph{J. Field Robot.},
  vol.~26, no.~4, pp. 358--390, 2009.

\bibitem{barnes2009autonomous}
D.~Barnes, S.~Pugh, and L.~Tyler, ``Autonomous science target identification
  and acquisition ({ASTIA}) for planetary exploration,'' in \emph{Proc. of
  IEEE/RSJ IROS}, 2009, pp. 3329--3335.

\bibitem{gallant2013rover}
M.~J. Gallant, A.~Ellery, and J.~A. Marshall, ``Rover-based autonomous science
  by probabilistic identification and evaluation,'' \emph{J. Intell. Robot.
  Syst.}, vol.~72, no. 3-4, p. 591, 2013.

\bibitem{pedersen2001autonomous}
L.~Pedersen, ``Autonomous characterization of unknown environments,'' in
  \emph{Proc. of IEEE ICRA}, 2001, pp. 277--284.

\bibitem{post2016planetary}
M.~A. Post, J.~Li, and B.~M. Quine, ``Planetary micro-rover operations on
  {M}ars using a {B}ayesian framework for inference and control,'' \emph{Acta
  Astronaut.}, vol. 120, pp. 295--314, 2016.

\bibitem{candelaIROS}
A.~Candela, D.~Thompson, E.~N. Dobrea, and D.~Wettergreen, ``Planetary robotic
  exploration driven by science hypotheses for geologic mapping,'' in
  \emph{Proc. of IEEE/RSJ IROS}, 2017.

\bibitem{brachman1985overview}
R.~J. Brachman and J.~G. Schmolze, ``An overview of the {KL-ONE} knowledge
  representation system,'' \emph{Cognitive Sci.}, vol.~9, no.~2, pp. 171--216,
  1985.

\bibitem{kononenko1993inductive}
I.~Kononenko, ``Inductive and {Bayesian} learning in medical diagnosis,''
  \emph{Appl. Artif. Intell.}, vol.~7, no.~4, pp. 317--337, 1993.

\bibitem{sowa2014principles}
J.~F. Sowa, \emph{Principles of semantic networks: Explorations in the
  representation of knowledge}.\hskip 1em plus 0.5em minus 0.4em\relax Morgan
  Kaufmann, 2014.

\bibitem{zhang2015mixed}
S.~Zhang, M.~Sridharan, and J.~L. Wyatt, ``Mixed logical inference and
  probabilistic planning for robots in unreliable worlds,'' \emph{IEEE Trans.
  Robot.}, vol.~31, no.~3, pp. 699--713, 2015.

\bibitem{hanheide2011exploiting}
M.~Hanheide, C.~Gretton, R.~Dearden, N.~Hawes, J.~Wyatt, A.~Pronobis,
  A.~Aydemir, M.~G{\"o}belbecker, and H.~Zender, ``Exploiting probabilistic
  knowledge under uncertain sensing for efficient robot behaviour,'' in
  \emph{Proc. of IJCAI}, 2011, pp. 2442--2449.

\bibitem{liedloff2013integrating}
A.~Liedloff, E.~Woodward, G.~Harrington, and S.~Jackson, ``Integrating
  indigenous ecological and scientific hydro-geological knowledge using a
  {Bayesian} network in the context of water resource development,'' \emph{J.
  Hydro.}, vol. 499, pp. 177--187, 2013.

\bibitem{castillo2012expert}
E.~Castillo, J.~M. Gutierrez, and A.~S. Hadi, \emph{Expert systems and
  probabilistic network models}.\hskip 1em plus 0.5em minus 0.4em\relax
  Springer Science \& Business Media, 2012.

\bibitem{apostolopoulos2001robotic}
D.~S. Apostolopoulos, L.~Pedersen, B.~N. Shamah, K.~Shillcutt, M.~D. Wagner,
  and W.~L. Whittaker, ``Robotic antarctic meteorite search: {O}utcomes,'' in
  \emph{Proc. of IEEE ICRA}, 2001, pp. 4174--4179.

\bibitem{sharif2015autonomous}
H.~Sharif, M.~Ralchenko, C.~Samson, and A.~Ellery, ``Autonomous rock
  classification using {B}ayesian image analysis for rover-based planetary
  exploration,'' \emph{Comput. Geosci.}, vol.~83, pp. 153--167, 2015.

\bibitem{beal2003variational}
M.~J. Beal, \emph{Variational algorithms for approximate {Bayesian}
  inference}.\hskip 1em plus 0.5em minus 0.4em\relax University of London
  United Kingdom, 2003.

\bibitem{murphy1999loopy}
K.~P. Murphy, Y.~Weiss, and M.~I. Jordan, ``Loopy belief propagation for
  approximate inference: An empirical study,'' in \emph{Proc. of Uncertain.
  Artif. Intel.}\hskip 1em plus 0.5em minus 0.4em\relax Morgan Kaufmann
  Publishers Inc., 1999, pp. 467--475.

\bibitem{krause2007near}
A.~Krause and C.~Guestrin, ``Near-optimal observation selection using
  submodular functions,'' in \emph{Proc. of AAAI}, 2007, pp. 1650--1654.

\bibitem{binney2012branch}
J.~Binney and G.~S. Sukhatme, ``Branch and bound for informative path
  planning.'' in \emph{Proc. of IEEE ICRA}, 2012, pp. 2147--2154.

\bibitem{bestprobabilistic}
G.~Best and R.~Fitch, ``Probabilistic maximum set cover with path constraints
  for informative path planning,'' in \emph{Proc. of ARAA ACRA}, 2016.

\bibitem{jawaid2015informative}
S.~T. Jawaid and S.~L. Smith, ``Informative path planning as a maximum
  traveling salesman problem with submodular rewards,'' \emph{Discrete Appl.
  Math.}, vol. 186, pp. 112--127, 2015.

\bibitem{wettergreen2014science}
D.~Wettergreen, G.~Foil, M.~Furlong, and D.~R. Thompson, ``Science autonomy for
  rover subsurface exploration of the {Atacama} desert,'' \emph{AI Mag.},
  vol.~35, no.~4, pp. 47--60, 2014.

\bibitem{girdhar2016modeling}
Y.~Girdhar and G.~Dudek, ``Modeling curiosity in a mobile robot for long-term
  autonomous exploration and monitoring,'' \emph{Auton. Robot.}, vol.~40,
  no.~7, pp. 1267--1278, 2016.

\bibitem{tabib2016efficient}
W.~Tabib, R.~Whittaker, and N.~Michael, ``Efficient multi-sensor exploration
  using dependent observations and conditional mutual information,'' in
  \emph{Proc. of IEEE SSRR}, 2016, pp. 42--47.

\bibitem{patten2017monte}
T.~Patten, W.~Martens, and R.~Fitch, ``{Monte Carlo} planning for active object
  classification,'' \emph{Auton. Robot.}, pp. 1--31, 2017.

\bibitem{browne2012survey}
C.~B. Browne, E.~Powley, D.~Whitehouse, S.~M. Lucas, P.~I. Cowling,
  P.~Rohlfshagen, S.~Tavener, D.~Perez, S.~Samothrakis, and S.~Colton, ``A
  survey of {Monte Carlo} tree search methods,'' \emph{IEEE Trans. Comp. Intel.
  AI}, vol.~4, no.~1, pp. 1--43, 2012.

\bibitem{yedidia2000generalized}
J.~S. Yedidia, W.~T. Freeman, Y.~Weiss \emph{et~al.}, ``Generalized belief
  propagation,'' in \emph{Adv. Neur. In.}, vol.~13, 2000, pp. 689--695.

\bibitem{kocsis2006bandit}
L.~Kocsis and C.~Szepesv{\'a}ri, ``Bandit-based {Monte Carlo} planning,'' in
  \emph{Proc. of Euro. Conf. Mach. Learn.}, 2006, pp. 282--293.

\bibitem{gelly2012grand}
S.~Gelly, L.~Kocsis, M.~Schoenauer, M.~Sebag, D.~Silver, C.~Szepesv{\'a}ri, and
  O.~Teytaud, ``The grand challenge of computer {Go}: {Monte Carlo} tree search
  and extensions,'' \emph{Commun. ACM}, vol.~55, no.~3, pp. 106--113, 2012.

\bibitem{heckerman1995learning}
D.~Heckerman, D.~Geiger, and D.~M. Chickering, ``Learning {B}ayesian networks:
  {T}he combination of knowledge and statistical data,'' \emph{Mach. Learn.},
  vol.~20, no.~3, pp. 197--243, 1995.

\bibitem{peynot2014learned}
T.~Peynot, S.-T. Lui, R.~McAllister, R.~Fitch, and S.~Sukkarieh, ``Learned
  stochastic mobility prediction for planning with control uncertainty on
  unstructured terrain,'' \emph{J. Field Robot.}, vol.~31, no.~6, pp. 969--995,
  2014.

\bibitem{thompson2007performance}
D.~R. Thompson and R.~Castano, ``Performance comparison of rock detection
  algorithms for autonomous planetary geology,'' in \emph{Proc. of IEEE Aero.
  Conf.}, 2007.

\bibitem{song2008automated}
Y.~Song and J.~Shan, ``Automated rock segmentation for {Mars} exploration rover
  imagery,'' in \emph{Proc. of Lunar and Planetary Science Conf.}, 2008.

\bibitem{dunlop2007multi}
H.~Dunlop, D.~R. Thompson, and D.~Wettergreen, ``Multi-scale features for
  detection and segmentation of rocks in {Mars} images,'' in \emph{Proc. of
  IEEE CVPR}, 2007.

\bibitem{achanta2012slic}
R.~Achanta, A.~Shaji, K.~Smith, A.~Lucchi, P.~Fua, and S.~S{\"u}sstrunk,
  ``{SLIC} superpixels compared to state-of-the-art superpixel methods,''
  \emph{IEEE Trans. Pattern Anal.}, vol.~34, no.~11, pp. 2274--2282, 2012.

\bibitem{potiris2014terrain}
S.~Potiris, A.~Tompkins, and A.~Goktogan, ``Terrain-based path planning and
  following for an experimental {Mars} rover,'' in \emph{Proc. of ARAA ACRA},
  2014.

\bibitem{blei2003latent}
D.~M. Blei, A.~Y. Ng, and M.~I. Jordan, ``Latent dirichlet allocation,''
  \emph{J. Mach. Learn. Res.}, vol.~3, no. Jan, pp. 993--1022, 2003.

\bibitem{choset2001coverage}
H.~Choset, ``Coverage for robotics--a survey of recent results,'' \emph{Ann.
  Math. Artif. Intel.}, vol.~31, no.~1, pp. 113--126, 2001.

\end{thebibliography}

\end{document}